\documentclass[nohyperref]{article}

\usepackage{microtype}
\usepackage{graphicx}
\usepackage{booktabs} %

\usepackage[hidelinks]{hyperref}       %

\usepackage[accepted]{icml2022}

\usepackage{amsmath}
\usepackage{amssymb}
\usepackage{mathtools}
\usepackage{amsthm}

\usepackage{url}            %
\usepackage{booktabs}       %
\usepackage{amsfonts}       %
\usepackage{nicefrac}       %
\usepackage{microtype}      %
\usepackage{xcolor}         %
\usepackage{graphicx,epstopdf}
\usepackage{adjustbox}
\usepackage{multirow}
\usepackage{csquotes}
\usepackage{wrapfig}
\usepackage{pifont}  %
\usepackage{makecell}  %

\usepackage{enumitem}
\usepackage{subcaption}
\usepackage{comment}

\usepackage{mathtools}

\usepackage{xspace}
\makeatletter
\DeclareRobustCommand\onedot{\futurelet\@let@token\@onedot}
\def\@onedot{\ifx\@let@token.\else.\null\fi\xspace}

\def\eg{\emph{e.g}\onedot} 
\def\ie{\emph{i.e}\onedot} 
 
\def\etc{\emph{etc}\onedot}

\makeatother

\usepackage{sistyle}
\SIthousandsep{,}

\usepackage{upgreek}

\def\mD{\mathcal{D}}

\def\mN{\mathcal{N}}

\def\1n{\mathbf{1}_n}
\def\0{\mathbf{0}}
\def\1{\mathbf{1}}

\def\R{{\mathbb R}}

\def\c{{\bf c}}

\def\f{{\bf f}}

\def\n{{\bf n}}

\def\x{{\bf x}}

\def\z{{\bf z}}

\DeclareMathOperator*{\bbE}{\mathbb{E}}

\newcommand{\cX}{\mathcal{X}}

\newcommand{\tavg}{\mbox{avg}}
\usepackage{physics}

\usepackage{array}
\newcolumntype{C}[1]{>{\centering\arraybackslash}m{#1}}  %

\newcommand{\secsym}{Sec\onedot}
\newcommand{\figsym}{Fig\onedot}
\newcommand{\equsym}{eq\onedot}

\newcommand{\secref}[1]{\secsym~\ref{#1}}
\newcommand{\figref}[1]{\figsym~\ref{#1}}
\newcommand{\equref}[1]{\equsym~\eqref{#1}}

\newcommand{\tb}[3]{{#1}^{#2}_{#3}}  %
\newcommand{\past}{{1..{t{-}1}}}  %

\usepackage{placeins}
\usepackage{xr}

\newcommand{\fT}{\mathcal{M}}
\newcommand{\fphi}{\phi}

\newcommand{\new}{}  %

\usepackage[capitalize,noabbrev]{cleveref}

\theoremstyle{plain}

\theoremstyle{definition}

\theoremstyle{remark}

\usepackage[textsize=tiny]{todonotes}

\icmltitlerunning{Style Equalization: Unsupervised Learning of Controllable Generative Sequence Models}

\begin{document}

\twocolumn[
\icmltitle{Style Equalization: Unsupervised Learning of Controllable \\ Generative Sequence Models }

\icmlsetsymbol{equal}{*}

\begin{icmlauthorlist}
\icmlauthor{Jen-Hao Rick Chang}{apple}  %
\icmlauthor{Ashish Shrivastava}{apple} %
\icmlauthor{Hema Swetha Koppula}{apple}
\icmlauthor{Xiaoshuai Zhang}{uscd}
\icmlauthor{Oncel Tuzel}{apple}
\icmlauthor{\normalfont \url{https://apple.github.io/ml-style-equalization}}{}
\end{icmlauthorlist}

\icmlaffiliation{apple}{Apple}
\icmlaffiliation{uscd}{University of California, San Diego}

\icmlcorrespondingauthor{Jen-Hao Rick Chang}{jenhao\_chang@apple.com}

\icmlkeywords{Style transform, generative model, controllable}

\vskip 0.3in
]

\printAffiliationsAndNotice{}  %

\begin{abstract}
Controllable generative sequence models with the capability to extract and replicate the style of specific examples enable many applications, including narrating audiobooks in different voices, auto-completing and auto-correcting written handwriting, and generating missing training samples for downstream recognition tasks.
However, under an unsupervised-style setting, typical training algorithms for controllable sequence generative models suffer from the training-inference mismatch, where the same sample is used as content and style input during training but unpaired samples are given during inference. 
In this paper, we tackle the training-inference mismatch encountered during unsupervised learning of controllable generative sequence models.
The proposed method is simple yet effective, where we use a style transformation module to transfer target style information into an unrelated style input. 
This method enables training using unpaired content and style samples and thereby mitigate the training-inference mismatch.
We apply style equalization to text-to-speech and text-to-handwriting synthesis on three datasets.  
We conduct thorough evaluation, including both quantitative and qualitative user studies.
Our results show that by mitigating the training-inference mismatch with the proposed style equalization, we achieve style replication scores comparable to real data in our user studies.

\end{abstract}

\section{Introduction}
\label{sec: intro}

\begin{figure*}[t]
	\centering
	\includegraphics[width=0.98\linewidth]{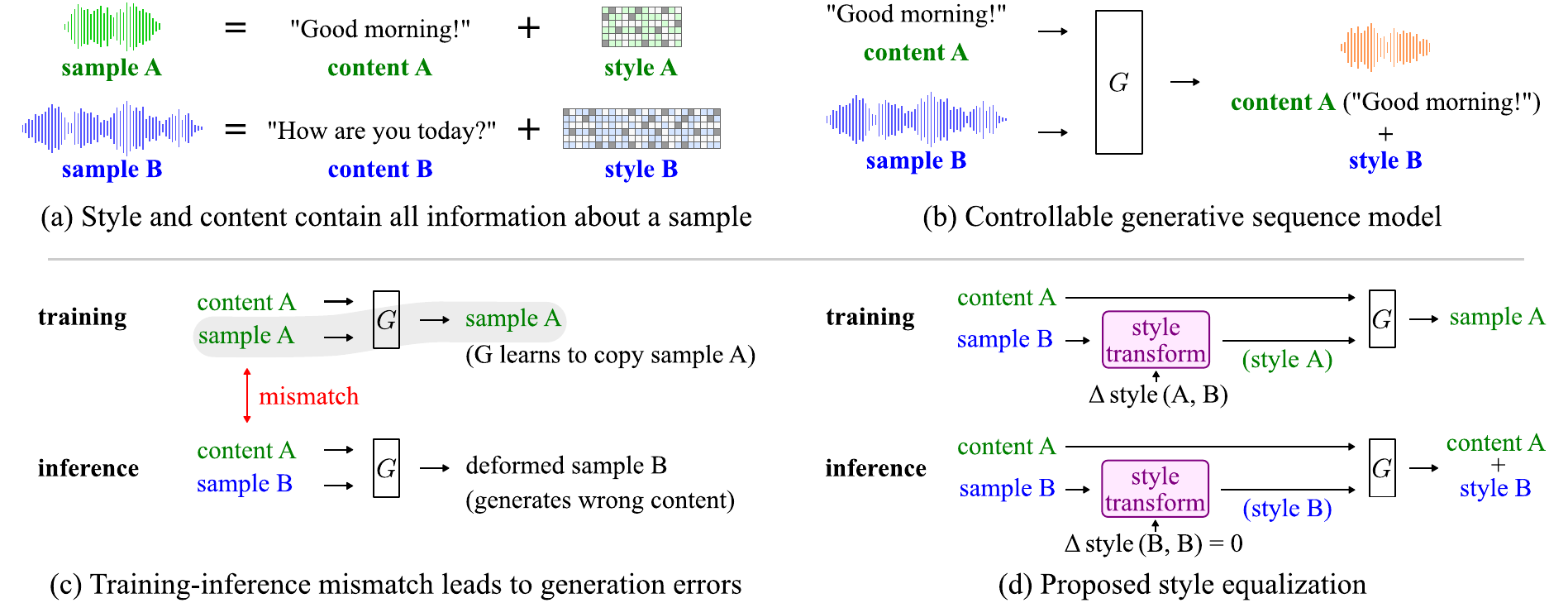}
	\caption{\textbf{Controllable generative models with sample-level style control.}  
		(a) The information contained in a sample can be divided into content (\ie, the text) and style (\ie, all other information besides content). 
		(b) Our goal during inference is to generate samples containing target content A in the style of sample B. Notice that sample B generally contains a different content. 
		(c) There exists a training-inference mismatch when learning these models in typical unsupervised training of controllable generative models. 
		During training, the \textit{same} sample is used as content input and style input, whereas during inference, content and style inputs are from different samples, \ie, the reference style sample contains a different content than the target content. The mismatch leads to incorrect content generation during inference. 
		(d) To mitigate the training-inference mismatch, the proposed style equalization takes unpaired samples as input during both training and inference. It transforms the style of sample B to that of A by estimating their style difference.  %
	}
	\label{figure: illustration}
\end{figure*}

The goal of controllable generative sequence models is to generate sequences containing target content in a target style. 
With the capability to select speaker voices, multi-speaker text-to-speech models have been successfully adopted in many voice assistants~\citep{arik2017deep,ping2018deep,hayashi2020espnet}. 
Many applications, however, require style controllability \textit{beyond} selecting speaker voices.
To perfectly reconstruct a speech example, we need to replicate not only the speaker's voice but also other aspects of style, including but not limited to prosody, intonation dynamics, background noise, echo, and microphone response appeared in the given sample.
To utilize synthetic data to analyze failures or biases of a downstream recognizer, we need a style representation that models the style distribution beyond speaker identity.
In these applications, style represents all information (except the content) to exactly reconstruct a sample, as illustrated in \figref{figure: illustration}a.
To model the temporal dynamics contained in input samples (\eg, speech and handwriting), our style representation is a time-varying sequence, instead of a fixed vector.
To capture the large variation of style, we learn the style representation in an \textit{unsupervised} manner from a reference sample, rather than using a few human-annotated attributes.

Our goal is to learn a controllable generative sequence model that controls its style with a reference example (\eg, an existing audio) and controls the content with a content sequence (\eg, text), as shown in \figref{figure: illustration}b.
Our training dataset $\cX$ is composed of $\left\{(\x^i, \c^i) \right\}_{i=1,\dots,n}$, where $\x^i {=} \left[\tb{x}{i}{1} \ \dots \ \tb{x}{i}{T_i} \, | \, \tb{x}{i}{t} \in \R^d \, \right]$ is the i-th sample and $\c^i {=} \left[\tb{c}{i}{1} \ \dots \ \tb{c}{i}{N_i} \, | \, \tb{c}{i}{j} \in \R^m \right]$ is the corresponding content sequence. 
In general, $\x^i$ and $\c^i$ have different lengths, \ie, $T_i \neq N_i$, and we do not assume to have the alignment between them.
For example, in text-to-speech synthesis, $\x^i$ is the mel-spectrogram of an audio sample, $\c^i$ is the corresponding phonemes of the spoken words.  %
While there exist methods, \eg, \cite{mcauliffe2017montreal,kotani2020generating}, that can segment phonemes from an mel-spectrogram or characters from a handwriting sequence, we choose not to rely on these information and directly learn the alignment using attention.
This allows the learned models to deal with ambiguous phoneme/character boundaries  (\eg, cursive handwriting).
We also do not assume to have any style supervision, including speaker or attribute labels, nor any grouping of the data based on style.

While the unsupervised setting requires only the essential information (\ie, samples and their content), it makes learning a controllable generative sequence model  difficult.
The main challenge is the mismatch between the inputs used during training and inference.
As shown in \figref{figure: illustration}c, during inference we pair arbitrary content A and reference sample B as inputs (\ie, nonparallel setting).  
However, due to the lack of ground truth containing content A and in the style of B, during training we pair content A and sample A (\ie, parallel setting).
In other words, we train the model under the parallel setting but we use the model in the non-parallel setting during inference.
Due to the training-inference mismatch, a well-performing model during training may perform poorly during inference.
If a generative model learns to utilize the content information in the style example, during inference the generative model will generate wrong content. %
This phenomenon is called \textit{content leakage}~\citep{hu2020unsupervised}.
In an extreme case, a model can learn to copy the reference sample to the output; despite its perfect training loss, it is useless because it always generates wrong content in practice.

This paper proposes a simple but effective technique to deal with the training-inference mismatch when we learn controllable auto-regressive models in an unsupervised-style manner.
As shown in \figref{figure: illustration}d, we train the model under the non-parallel setting, \ie, we pair arbitrary content A with an arbitrary sample B from the training dataset.
Instead of directly using sample B as style (in which case we have no ground-truth output), we jointly learn a style transformation function, which estimates the style difference between A and B and transforms the style of sample B to the style of A.
The generative model then takes content A and the transformation output (that contains the style of A and the unrelated content information from B) to reconstruct sample A.
The proposed method enables us to use sample A as the ground truth while learning in the non-parallel setting---the intended usage during inference. \new
Additionally, our method provides a systematic way to interpolate between the style of two samples by scaling the estimated style difference between two reference samples.
We call the method \textit{style equalization}.
Note that for style equalization to work, the style transformation and difference estimator need to be carefully designed, such that no content information from content A can be transferred through sample B.  
We defer the discussion to~\secref{sec: style equalization}.

We apply the proposed method on two signal domains, speech and online-handwriting, and evaluate the performance carefully via quantitative evaluation (by computing content error rates) and conducting qualitative user studies. 
Experimental results show that by tackling training-inference mismatch with the proposed style equalization, we are able to learn strong unsupervised controllable sequence generative models that have competitive performance even when compared to existing methods that utilize style supervision like speaker labels.
On LibriTTS, style equalization achieves close style replication (3.5 real oracle vs. 3.5 proposed in style opinion score) and content reproduction errors (6.6\% real oracle vs. 9.5\% proposed) to real samples.

\section{Related Work}
\label{sec: related_work}

Controllable generative sequence models are not new in the literature; however, the majority of these methods utilize style supervision, whereas the paper focuses on developing an unsupervised-style method.
Table~\ref{table: related work} provides an overview of the related works.

\paragraph{Unsupervised-style  methods.}

Unsupervised methods extract style information directly from samples, \ie, without any style labels or pretrained style embeddings. 
Existing unsupervised methods train models under the parallel setting, as shown in \figref{figure: illustration}c. 
To prevent content leakage, most existing methods introduce a bottleneck on the capacity of the style encoder by representing style as a single (time-invariant) vector and limiting its dimension~\citep{wang2018style,hsu2018hierarchical,hu2020unsupervised,ma2018neural}. 
\citet{wang2018style} propose Global Style Token (GST), which represents a style vector as a linear combination of a learned dictionary (called style tokens) shared across the dataset. 
The number of style tokens (the implicit dimension of the style vector) is carefully controlled to prevent content leakage. 
As we will see in \secref{sec: background}, the bottleneck not only reduces the amount of content information contained in the style vector but also  sacrifices  style information.

Alternative loss formulations have also been proposed to limit content information contained in the style representation.  
\citet{hu2020unsupervised} minimize the mutual information between the style vector and the content sequence but requires a pretrained content encoder and adversarial learning, which makes training their model difficult.
\citet{hsu2018hierarchical} approximate the posterior distribution of the style vector using a mixture of Gaussian distributions with a small number of mixtures.
\citet{ma2018neural} utilize a discriminator conditioned on both the generated output and the content (similar to a content recognizer).
\citet{akuzawa2018expressive} anneal the Kullback-Leibler divergence to control the amount of information contained in style.
\citet{henter2018deep} utilize phoneme segmentation~\citep{mcauliffe2017montreal} to avoid learning the alignment between content $\c$ and output $\x$.

Priming is a technique that is introduced to control the style of auto-regressive generative sequence models~\citep{graves2013generating,aksan2018deepwriting}. 
Since the hidden state of a Recurrent Neural Network (RNN) contain all information about current generation, including style, we can initialize the RNN by pre-rolling the reference sample through the RNN.
Utilizing priming requires the content of the reference style. 
For example, \citet{aksan2018deepwriting} learn a character recognizer and use it during inference. 
Moreover, since the hidden state contains residual content from the reference example, it often generates unexpected artifacts at the beginning of the sequence, as will be seen in \secref{sec: experiments}.

\paragraph{Supervised-style methods.}

Many existing controllable generative models utilize style supervision, either directly by passing attribute labels as inputs or implicitly by grouping training data with their attribute labels. 
In the following, we briefly introduce various supervised controllable sequence models.
While using style supervision avoids training-inference mismatch, it limits the style control on a few sparsely-defined attribute classes.  
For instance, given a speech audio, we can recognize the spoken texts, the accent, or even the speaker, but provided solely with these attribute labels, it is impossible to exactly reconstruct the original speech audio.
The sparsely-defined attributes are insufficient to capture the entire style information.

User identifications or their embeddings have been used to learn multi-speaker text-to-speech models~\citep{jia2018transfer,arik2017deep, kameoka2020many,donahue2020end,chen2021adaspeech,dhariwal2020jukebox,valle2020mellotron,kim2020glow,hayashi2020espnet,sun2020fully,skerry2018towards,sun2020generating}, voice conversion models~\citep{qian2019autovc,lee2021voicemixer} and handwriting models~\citep{kotani2020generating,bhunia2021handwriting,kang2020ganwriting,davis2020text}.
In addition to user identifications, predefined features like pitch, phoneme duration, loudness, and timbre have also been used by existing methods~\citep{ren2020fastspeech, qian2020unsupervised, dhariwal2020jukebox, valle2020mellotron}.
Instead of using speaker labels as input, \citet{kameoka2018stargan,kaneko2018cyclegan,kaneko2019cyclegan,kaneko2019stargan} group training samples by their speaker labels and apply adversarial learning to learn voice conversion models that change speaker voices while keeping the content of the input.

\paragraph{Image methods.}

Controllable generative models have also been developed for images~\citep{HarkonenGANSpace2020,esser2019unsupervised,singh_cvpr2019,NIPS2017_faderNet, Karras_stylegan_v2_2020,brock2018large,Collins_2020_CVPR,Shen_2020_CVPR,Esser_2020_CVPR,GANalyze_GoetschalckxAOI19,pavllo2020controlling,zhang2018separating,chan2021pi,kwon2021diagonal,kazemi2019style}, which control the object class, pose, lighting, \etc, of an image.
Many image style transform methods have also been developed~\citep{isola2017image,zhu2017unpaired,gatys2016neural,kotovenko2019content}.
However, there is a fundamental difference between image and sequence problems.
In image generative models, we do not need to learn the content-output alignment.  
The content is usually defined globally as an image class or as pixel labels, \eg, segmentation map. 
In contrast, our content is given as text, the output is mel-spectrogram of a waveform, and the content and output have different lengths.
To utilize the input content sequence, generative sequence models need to align the content and the output sequences and translate text to the output signal modality.  
The complication exacerbates the training-inference mismatch for sequence methods, since copying the style input is easier than utilizing the input content.

\section{Controllable Generative Sequence Models}
\label{sec: background}

We focus on learning controllable auto-regressive generative models,  $p(x_t | z_t, \x_\past, \c)$, where $\x = [x_1, \dots, x_T ]$ is the output sequence, $\c$ is the content sequence, and $\z = [z_1, \dots, z_T \, | \, z_t \in \R^\ell]$ is the reference style information. 
Note that, in our model, style is also represented as a sequence that changes over time. 
Under the style-unsupervised setting, we are given a dataset $\cX = \{(\x^i, \c^i), i \in \{1 \dots n\}\}$ that contains the ground-truth output sequence $\x^i$ and the corresponding content $\c^i$, but we do not have supervision on $\z$. 
Therefore, we treat $z_t$ as a latent variable with a learnable prior distribution $p(z_t | \x_\past, \c)$ and optimize the log-likelihood of $\x$ conditioned on $\c$,  $\bbE_{(\x, \c)} \log p(\x | \c)$.

Our model is a variational RNN~\citep{chung2015recurrent}, and we maximize a variational lower bound of the likelihood:  
\begin{align}
	& \bbE_{(\x, \c)} \ \log p(\x | \c)  = \bbE_{(\x, \c)} \sum_{t=1}^{T} \log p(x_t | \x_\past, \c) \nonumber  \\
	& = \bbE_{(\x, \c)} \sum_{t=1}^{T} \log \bbE_{z_t \sim p(z_t | \x_\past, \c)} p(x_t | z_t, \x_\past, \c)  \nonumber \\
	& \ge \bbE_{(\x, \c)}  \sum_{t=1}^{T} \bbE_{z_t \sim q(z_t|\x, \c)} \log p(x_t | z_t, \x_\past, \c) \nonumber \\
	& \qquad \qquad  - \mD_{KL} \left( q(z_t | \x, \c) \, || \, p(z_t | \x_\past, \c) \right),  	\label{eq: lower bound}
\end{align}
where $\mD_{KL}$ represents the Kullback-Leibler (KL)-divergence.  
In \equref{eq: lower bound}, we use the chain rule to expand $p(\x | \c)$ into $p(x_1|\c) \, p(x_2 | x_1, \c) \cdots p(x_T | \x_{1..T{-}1}, \c)$, introduce the variational approximation $q(z_t| \x, \c)$ of the posterior distribution $p(z_t | \x, \c)$ for all $t$, and apply Jensen's inequality. 
Note that since $q$ is a variation approximation of the posterior distribution, it can be conditioned on any variable.
As we will see in \secref{sec: style equalization}, the proposed style equalization manipulates the input to $q$ such that  we condition $q$ on a style-transformed unrelated sample $\x'$ instead of $\x$.

\begin{figure*}[t]
	\centering
	\includegraphics[width=0.92\linewidth]{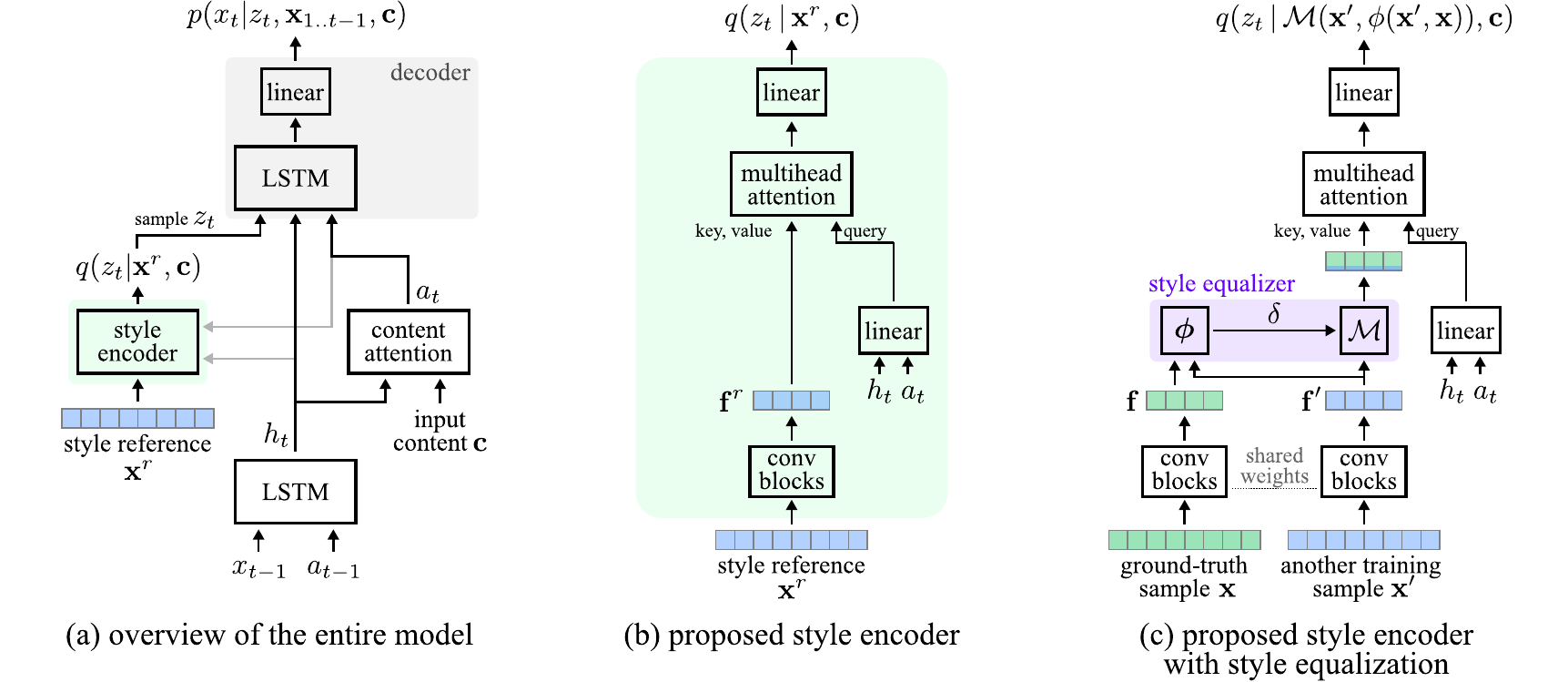}
	\vspace{-1mm}
	\caption{\textbf{Model used in the paper.} 
		(a) an overview of the entire model, which includes a style encoder (in green), content attention, and decoder (in gray). Note that the input content $\c$ can be the output of a content-embedding network (used in speech synthesis) or one-hot encoding of characters (used in handwriting synthesis). $a_t$ is the output of the content attention at time $t$, which is a linear combination of the elements in $\c$.
		(b) the proposed style encoder without style equalization. 
		(c) the proposed style encoder with the style equalization. $\fphi$ computes the vector $\delta$ that encodes the style difference between $\x'$ and $\x$. $\fT$ applies this transformation to $\x'$ to match the style of $\x$.}
	\label{figure: arch}
\end{figure*}

\figref{figure: arch}a shows an overview of the network used in the paper --- the input content $\c$ is processed by the content attention, the style encoder (shown in green) models $q(z_t | \x^r, \c)$, and the decoder (shown in gray) models $p(x_t | z_t, \x_\past, \c)$ using output from the style encoder and the content attention.
Note that during inference, $\x^r$ is the reference example that we replicate the style of, and it generally contains an unrelated content, \ie,  $\c^r \neq \c$.
During training, the ground-truth sample $\x$ is used as the reference style to optimize \equref{eq: lower bound}, \ie, $\x^r = \x$ and $\c^r = \c$, which is different from inference.
Therefore a generative model can learn to copy the style input to the output (and ignore the content input), leading to incorrect generation during inference.
This phenomenon can be remedied by limiting the capacity of the style encoder, \eg, by decreasing the dimensionality of the style representation.
However, to achieve the lower bound of \equref{eq: lower bound} and hence a higher generation quality, we need the style encoder to contain enough capacity such that $q(z_t| \x, \c) \equiv p(z_t | \x, \c)$ for all $t$. 
In this paper, we provide an alternative training procedure that bypasses this trade-off, allowing us to use a powerful style encoder (shown in \figref{figure: arch}b) with a high-dimensional style representation.  %

\section{Style Equalization}
\label{sec: style equalization}

Let $(\x, \c)$ and $(\x', \c')$ be two samples from the training set. 
To match the inference setting, we should train the model using $\c$ as content and $\x'$ as style input. 
However, neither $\x$ nor $\x'$ can be used as the ground-truth output sequence. 
If we use $\x$ as the ground-truth output sequence but $\x'$ as style input, the generative model will learn to ignore the style encoder since $\x'$ contains unrelated style information. 
In other words, the variational approximation $q(\z | \x', \c)$ is a poor approximation to the true posterior $p(\z | \x, \c)$. 
Alternatively, if we use $\x'$ as the ground-truth output sequence, the content given in $\c$ will be ignored.

We introduce a learnable style transformation function $\fT(\x', \delta)$  to transform the style of $\x'$ by the amount specified by a style difference vector $\delta \in \R^k$. 
We first estimate $\delta$ with a learnable function $\fphi$, \ie $\delta = \fphi(\x', \x)$;  we then transform $\x'$ with $\fT(\x', \delta)$. 
By jointly optimizing $\fT$, $\fphi$, and the rest of the generative model using \equref{eq: lower bound} (with the input to $q$ modified as described), the model learns to transfer style information from $\x$ through $\fT$ to maximize the log-likelihood of the ground-truth $\x$. 
In other words, we approximate the posterior distribution $p(\z | \x, \c)$ with $q(\z | \fT(\x',  \fphi(\x', \x)), \c)$, which is a better approximation than $q(\z | \x', \c)$. 
We call this method \textit{style equalization}.

Note that, for style equalization to be successful, $\fT \circ \fphi$ should not transfer any content-related information (\eg, copy the entire sequence) from $\x$ but only its style information so that the decoder will utilize the transferred style and will rely on provided content input to generate the output. 
Therefore the design of $\fT$ is critical.

\paragraph{Design of $\fT$ and $\fphi$.}
\label{sec: choice of T and phi}

An important observation we use in the design of $\fT$ and $\fphi$ is that content information (\ie, sequence of phonemes or characters) is strongly time-dependent whereas the style can be reasonably well approximated by a time-independent representation (\eg, voice characteristics of a speaker and microphone response, \etc).
By designing $\fphi$ such that no time-dependent information is stored in $\delta$, we can satisfy that content-related information is not leaked but still transfer the time-independent style information.
Thus, we utilize average pooling over the time dimension in $\fphi$, and to improve the time-invariant property of the convolutional network in our style encoder, we use convolutional filters without padding and with low-pass filtering before down-sampling~\citep{zhang2019making}. 

As shown in \figref{figure: arch}c, to estimate the style difference vector $\delta$ between two sequences $\x$ and $\x'$, we first compute their style features $\f$ and $\f'$ using a convolutional network. 
Note that $\f$ and $\f'$ are $s$-dimensional feature sequences with different lengths.
We define
\begin{align}
	& \fphi(\x', \x) = \tavg(A \, \f) - \tavg(A \, \f')  \mbox{ and } \nonumber  \\
	& \fT(\x', \fphi(\x', \x)) = \f' + A^\top \fphi(\x', \x),
	\label{eq: T and phi}
\end{align}
where $\tavg$ represents taking mean across time, $A \in \R^{k \times s}$ is a learnable linear transform, and $\f' + A^\top \fphi(\x', \x)$ means that the vector $A^\top \fphi(\x', \x) {\in} \R^s$ is added to each time step of $\f'$.
Intuitively, the design assumes that the style information lies on a $k$-dimensional subspace, and we equalize the style between $\x'$ and $\x$ by minimizing their differences in the subspace. 
It also satisfies the identity property by construction --- $\fphi(\x^r, \x^r) = 0$ and $\fT(\x^r, 0) = \f^r$ --- which enables style interpolation (see below) and allows us to treat style equalization as a training procedure and remove it from the model during inference (as shown in \figref{figure: arch}b).
Note that making a convolutional network entirely time-invariant is an important ongoing research problem~\cite{karras2021alias}; please see \secref{sec: limitations} for the discussions on our limitations.

\paragraph{Interpolation between two styles.}
Once learned, $\fT$ and $\fphi$ can be used to interpolate style during inference.
Given two style references, $\x^s$ and $\x^t$, we interpolate between them with $\fT(\x^s, \alpha \, \fphi(\x^s, \x^t))$, where a scalar $\alpha \, {\in} \, \R$ controls the interpolation.
By changing $\alpha$, we traverse a one-dimensional manifold that starts from the original style (with $\alpha = 0$, since $\fT(\x^s,  \fphi(\x^s, \x^s))= \f^s$) and ends at the target style (with $\alpha = 1$). 
Note that, unlike existing generative models that support style interpolation in post-processing, $\fT$ and $\fphi$ are trained to transform style by design.

\section{Experiments}
\label{sec: experiments}

To demonstrate the generality of the proposed method, we train and \new evaluate it on two signal domains, speech and handwriting, with the same model architecture design. 
In the following, we introduce the model architecture used in the experiments, the baselines, the metrics, and the results. 
More details are provided in \secref{sec: model architecture details}.

\subsection{Model Architecture}
\label{sec: model architecture: handwriting}

Our model is auto-regressive and composed of (i) a decoder that is modeling $p(x_t | z_t, \x_\past, \c)$, (ii) a content attention module, (iii) a style encoder that is modeling $q(z_t | \cdot, \c)$, and (iv) a network that models the prior distribution of $z_t$. 
\figref{figure: arch}a shows an overview of the model.
The backbone of the model, namely the content attention and the decoder, uses a standard architecture that was proposed by~\citet{graves2013generating} for handwriting synthesis and later extended to speech in variations of Tacotron~\citep{shen2018natural,wang2018style}.
The variational approximation $q(z_t | \cdot)$ and the prior distribution are modeled as multivariate Gaussian distributions with a diagonal covariance matrix.

Our style encoder is composed of a convolutional network and a multi-head attention layer, as shown in \figref{figure: arch}b.
The convolutional network extracts the style feature sequence $\f^r$ from the reference style input $\x^r$.
We use multi-head attention to extract relevant style information at every time step from $\f^r$ with the query computed from the hidden state of the LSTM, $h_t$, which contains information about past generations, and the currently focused content $a_t$.
Thus, while $\delta$ contains only time-invariant information, our style representation is a time-varying sequence.
The intuition is if the model utters a particular phoneme, it should be able to find the information in the style reference and mimic it.

For style equalization, we insert $\fT$ and $\fphi$ into the style encoder, as shown in \figref{figure: arch}c.  
Since style equalization is only able to transfer time-independent information, when we utilize this procedure, the network will not be able to learn time-dependent style. 
To enable learning time-dependent style information during training, half of the batches, we use $\x' = \x$, which means that the difference vector $\delta = 0$, hence the decoder directly uses the ground-truth style information which contains time-dependent style information. 
We analyze the effect of style attention and its ability to represent time-varying style information in \secref{sec: style bottleneck}.

\subsection{Speech Synthesis}

We train and evaluate the proposed method on two multi-speaker speech datasets. 
VCTK dataset \citep{yamagishi2019vctk} contains 110 speakers and 44 hours of speech, and LibriTTS dataset~\citep{zen2019libritts} contains 2,311 speakers and 555 hours of speech in the training set.  %

\paragraph{Baselines.} 

We compare the proposed method with Global Style Tokens (GST-n)~\citep{wang2018style} with various numbers of tokens n. 
For completeness, we also compare with Tacotron 2~\citep{shen2018natural} (even though it does not have style control), Tacotron-S and GST-nS.  
Tacotron-S / GST-nS are Tacotron / GST-n with style supervision --- a pretrained speaker embedding \citep{snyder2018x} that was trained on the VoxCeleb dataset, which contains $2,000$ hours of speech from $7,000$ speakers~\citep{chung2018voxceleb2}. 
We use ESPnet-TTS \citep{hayashi2020espnet}, a widely used implementation of the baselines and follow their training recipe. 
They achieve similar performance as those listed in the original papers.
With no publicly available implementation, we cannot compare with \citep{hsu2018hierarchical} and \citep{ma2018neural}.
All methods are trained on the same dataset, while Tacotron-S and GST-nS have additional style supervision from VoxCeleb dataset.
All methods output $80$-dimensional mel-spectrograms with the sampling rate equal to 22,050 Hz and the window size equal to 1,024, which are converted to waveforms using a pretrained WaveGlow vocoder~\citep{prenger2019waveglow}. 
The content input is represented as phonemes, following \citep{shen2018natural}.

\paragraph{Metrics.}
\label{sec: speech metric}

We measure the content generation errors as Word Error Rate (WER), using a pretrained speech recognition model, ESPnet~\citep{kamo_naoyuki_2021_4604066}.
To evaluate the style replication accuracy of the methods, we use a speaker classification network \citep{deng2019arcface} (see \secref{sec: style classifier network}) and measure the style similarity between reference and output generations.
We report \texttt{cos-sim},  which is the cosine similarity between the reference example and its corresponding output generation,  and \texttt{sRank}, which is the rank of the reference speaker out of all speakers based on their cosine similarities. 
We report the empirical mean and standard deviation of these metrics on 100 pairs of reference and synthetic samples.

We also report the style opinion score following the protocol used by \citet{zhao2020voice}.
To evaluate the style similarity between a generated output and a style reference, users were given pairs of reference and synthesized audio, and asked if ``the two samples could have been produced by the same speaker in a similar environmental condition", and asked to score with `` 4 (Absolutely same)", ``3 (Likely same)", ``2 (Likely different)", ``1 (Absolutely different)". 
We synthesized 100 samples using each method with the same style example and target content.  
A total of $15$ users participated in the study, and we collected $630$ responses. 

We also provide an oracle (a pseudo upper-bound) where we select a different real speech sample from the same speaker from the dataset, and evaluate style similarity and content error. 
This provides a good calibration for our evaluation metrics and opinion studies.

\begin{table}[t]
	\caption{Quantitative results on VCTK dataset. The reference style inputs are seen (randomly selected from the training set).  
	WER measures content accuracy; cosine-similarity (cos-sim) and sRank measure style similarity.   
	}
	\label{table: vctk}
	\def\tablewidth{18mm} 
	\centering
	\setlength{\tabcolsep}{3pt}
	\begin{adjustbox}{max width=\linewidth}
	\begin{tabular}{p{16mm}rrrrrr}
		\toprule
		\multirow{2}[3]{*}{Method} & 
		\multicolumn{3}{c}{Parallel text}  &   
		\multicolumn{3}{c}{Nonparallel text}  \\ 
		\cmidrule(lr){2-4} \cmidrule(lr){5-7}  %
		& 
		\multicolumn{1}{c}{WER (\%)}  & \multicolumn{1}{c}{cos-sim $\uparrow$} & \multicolumn{1}{c}{sRank $\downarrow$} &  
		\multicolumn{1}{c}{WER (\%)}  & \multicolumn{1}{c}{cos-sim $\uparrow$} & \multicolumn{1}{c}{sRank $\downarrow$}  \\
		\midrule
		Tacotron & %
		$16.0 \pm 1.7$ & $0.05 \pm 0.13$  & $53.1 \pm 29.1$  & 
		$16.4 \pm 1.2$ & $0.05 \pm 0.12$ & $53.9 \pm 27.8$  \\
		Tacotron-S & %
		$13.6 \pm 0.7$ & $0.24 \pm 0.18$  & $16.4 \pm 20.9$  & 
		$16.3 \pm 0.4$ & $0.22 \pm 0.18$ & $18.0 \pm 21.9$  \\
		GST-16 & %
		$18.6 \pm 0.9$ &  $0.23 \pm 0.15$ & $21.4 \pm 21.9$  & 
		$18.5 \pm 1.1$ & $0.23 \pm 0.16$ & $21.1 \pm 22.4$   \\
		GST-64 & %
		$16.9 \pm 0.5$  & $0.23 \pm 0.17$  & $24.4 \pm 23.1$  & 
		$27.5 \pm 0.4 $ & $0.22 \pm 0.16$ & $25.2 \pm 24.4$  \\
		GST-16S & %
		$8.3 \pm 0.1$ & $0.34 \pm 0.18$ & $10.8 \pm 15.2$ & 
		$17.7 \pm 0.8$ & $0.31 \pm 0.17$ & $13.0 \pm 20.0$  \\
		GST-64S & %
		$14.1 \pm 0.3$ & $0.33 \pm 0.18$ & $11.4 \pm 16.3$ &
		 $24.7 \pm 1.0$ & $0.32 \pm 0.18$ & $12.7 \pm 18.1$  \\ %
        \midrule
		Proposed & 
		$\mathbf{7.4 \pm 0.2}$ & $\mathbf{0.73 \pm 0.12}$ & $\mathbf{1.5 \pm 2.1}$  & 
		$\mathbf{9.5 \pm 0.4}$ & $\mathbf{0.64 \pm 0.14}$ & $\mathbf{1.9 \pm 4.2}$  \\
        \midrule
		Oracle & %
		$6.6 \pm 0.0$   & $1.0 \pm 0.0$  & $1.0 \pm 0.0$ & 
		$6.6 \pm 0.0$   & $0.57 \pm 0.16$ & $1.6 \pm 4.1$  \\
		\bottomrule
	\end{tabular}
	\end{adjustbox}
	\vspace{-1mm}
\end{table}

\begin{table*}[t]
	\caption{{Quantitative results on LibriTTS-all-960 dataset.}
	}
	\vspace{-3mm}
	\label{table: libritts}
	\def\tablewidth{16mm} 
	\centering
	\setlength{\tabcolsep}{3pt}
	\begin{adjustbox}{max width=0.88\linewidth}
		\begin{tabular}{p{16mm}rrrrrrrrr}
			\toprule
			\multirow{2}[3]{*}{Method} & \multicolumn{3}{c}{Seen speakers, parallel text}  &   \multicolumn{3}{c}{Seen speakers, nonparallel text}    &  \multicolumn{3}{c}{Unseen speakers, nonparallel text}   \\ 
			\cmidrule(lr){2-4} \cmidrule(lr){5-7} \cmidrule(lr){8-10}     
			& 
			\multicolumn{1}{c}{WER (\%)}  & \multicolumn{1}{c}{cos-sim $\uparrow$} & \multicolumn{1}{c}{sRank $\downarrow$} &  
			\multicolumn{1}{c}{WER (\%)}  & \multicolumn{1}{c}{cos-sim $\uparrow$} & \multicolumn{1}{c}{sRank $\downarrow$} &  
			\multicolumn{1}{c}{WER (\%)}  & \multicolumn{1}{c}{cos-sim $\uparrow$} & \multicolumn{1}{c}{sRank $\downarrow$} \\
			\midrule
			Tacotron & %
			$64.4 \pm 4.1$ & $0.00 \pm 0.10$  & $1218 \pm 671$  & 
			$52.2 \pm 1.7$ & $0.01 \pm 0.10$ & $1140 \pm 651$  & 
			$52.2 \pm 0.6$ & $0.00 \pm 0.10$ & $847 \pm 563$  
			\\
			Tacotron-S & %
			$14.9 \pm 0.0$ & $0.23 \pm 0.22$  & $430 \pm 584$  & 
			$18.7 \pm 0.2$ & $0.24 \pm 0.22$ & $370 \pm 562$  & 
			$13.5 \pm 0.0$ & $0.16 \pm 0.16$ & $289 \pm 436$  
			\\
			GST-64 & %
			$38.2 \pm 2.2$ &  $0.12 \pm 0.19$ & $706 \pm 701$  & 
			$33.3 \pm 3.4$ & $0.12 \pm 0.19$ & $700 \pm 739$  &
			$30.4 \pm 2.2$ & $0.09 \pm 0.16$ & $535 \pm 610$   
			\\
			GST-192 & %
			$19.3 \pm 0.3$  & $0.10 \pm 0.17$  & $786 \pm 725$  & 
			$17.8 \pm 0.6$ & $0.09 \pm 0.16$ & $823 \pm 719$  & 
			$18.0 \pm 0.7$ & $0.07 \pm 0.14$ & $587 \pm 582$  
			\\
			GST-64S & %
			$19.7 \pm 0.7$ &  $0.39 \pm 0.23$ & $150 \pm 305$  & 
			$20.4 \pm 1.6$ & $0.40 \pm 0.23$ & $143 \pm 334$   &
			$16.5 \pm 0.2$ & $0.28 \pm 0.17$ & $121  \pm 259$   
			\\
			GST-192S & %
			$13.8 \pm 0.7$ & $0.39 \pm 0.23$ & $137 \pm 309$ &
			$15.4 \pm 1.1$ & $0.41 \pm 0.22$ & $126 \pm 317$   &
			$13.4 \pm 0.2$ & $0.29 \pm 0.18$ & $139 \pm 316$   
			\\
			\midrule
			Proposed & 
			$\mathbf{6.2 \pm 0.5}$ & $\mathbf{0.82 \pm 0.14}$ & $\mathbf{1.7 \pm 4.1}$  & 
			$\mathbf{9.4 \pm 0.3}$ & $\mathbf{0.78 \pm 0.14}$ & $\mathbf{1.8 \pm 6.0}$  &
			$\mathbf{7.6 \pm 0.9 }$ & $\mathbf{0.57 \pm 0.15}$ & $\mathbf{7.4\pm 42.6}$  
			\\
			\midrule
			Oracle & %
			$6.5 \pm 0.0$   & $1.0 \pm 0.0$   & $1.0 \pm 0.0$  & 
			$6.5 \pm 0.0$ & $0.85 \pm 0.06$ & $1.0 \pm 0.0$  & 
			$6.5 \pm 0.0$ & $0.50 \pm 0.23$ & $3.6 \pm 24.9$  
			\\
			\bottomrule
		\end{tabular}
	\end{adjustbox}
\end{table*}

\begin{table*}[t]
	\caption{{Style opinion scores of speech synthesizers.} 
	}
	\vspace{-3mm}
	\label{table: speech qualitative}
	\def\tablewidth{18mm} 
	\centering
	\setlength{\tabcolsep}{3pt}
	\begin{adjustbox}{max width=0.88\linewidth}
		\begin{tabular}{rrrrrrrrrrrr}
			\toprule
			\multicolumn{4}{c}{\textbf{VCTK, seen speakers}}  &   
			\multicolumn{4}{c}{\textbf{LibriTTS, seen speakers}}    & 
			\multicolumn{4}{c}{\textbf{LibriTTS, unseen speakers}}   \\ 
			\cmidrule(lr){1-4} \cmidrule(lr){5-8} \cmidrule(lr){9-12}     
			\multicolumn{1}{c}{GST-64} & 
			\multicolumn{1}{c}{GST-16S} &  
			\multicolumn{1}{c}{Proposed} &  
			\multicolumn{1}{c}{Oracle}  & %
			\multicolumn{1}{c}{GST-192} & 
			\multicolumn{1}{c}{GST-192S} &  
			\multicolumn{1}{c}{Proposed} &  
			\multicolumn{1}{c}{Oracle}  & %
			\multicolumn{1}{c}{GST-192} & 
			\multicolumn{1}{c}{GST-192S} &  
			\multicolumn{1}{c}{Proposed} &
			\multicolumn{1}{c}{Oracle}   %
			\\
			\midrule
			$2.1 \pm 1.0$  & %
			$3.3 \pm 0.9$  & %
			$3.8 \pm 0.4$  & %
			$3.8 \pm 0.4$  & %
			$1.4 \pm 0.6$  & %
			$2.8 \pm 1.0$  & %
			$\mathbf{3.6 \pm 0.6}$  & %
			$3.5 \pm 0.9 $  & %
			$1.2 \pm 0.5$  & %
			$2.6 \pm 0.9$  & %
			$\mathbf{3.5 \pm 0.7}$  & %
			$3.5 \pm 0.9$   %
			\\
			\bottomrule
		\end{tabular}
	\end{adjustbox}
	\vspace{-3mm}
\end{table*}

\paragraph{Results.}

Table~\ref{table: vctk} shows the results on VCTK dataset.
Let us first look at the parallel-text setting. %
Without any style control, Tacotron achieves a low cosine similarity. 
Utilizing GST and speaker embedding (Tacotron-S and GST-nS) improve cosine similarities.
With the proposed style equalization and the style encoder, our model achieves the highest similarity and  lowest sRank.

When comparing the WERs between the parallel and non-parallel settings, we can see the adversarial effect of content leakage --- the models with a high-capacity style encoder (GST-64, GST-16S, and GST-64S) produce a much higher word-error rate in the non-parallel setting than in the parallel setting, while the small capacity GST-16 is largely unaffected. 
In comparison, our model is less affected by the parallel / nonparallel settings. 
The supplemental website shows the generated audios.

Table~\ref{table: libritts} shows the results on LibriTTS dataset.
As can be seen from the results, the large variety (\eg, more accents, higher background noise, difficult microphone effects)  makes it a more difficult dataset than VCTK, as reflected by the higher WERs of Tacotron than those in VCTK.
The proposed method, on the other hand, is able to learn from the noisy data and mimic the wide variety of styles in the dataset.  %
In addition, since the training set contains more than $2,000$ speakers, the proposed method learns to generalize and mimic the voices of unseen speakers in the validation set, as demonstrated by the high cosine similarity in the right-most column of Table~\ref{table: libritts}. 
The supplemental website shows the generated audios, including those by a model using the same architecture as ours but trained without style equalization. 
While the model performs well in the parallel setting, it fails to generate the correct content in the nonparallel setting, and we can clearly hear the content from the nonparallel style input leaked into the generated output. 
In comparison, utilizing style equalization during training significantly improves the problem.
Please also see the supplemental website for results with style sampled from the prior distribution or interpolated between two examples.

Table~\ref{table: speech qualitative} shows style opinion scores on the two datasets. 
The proposed method achieves high scores, similar to those of the oracle.

Finally, it is worth noting that by directly mimicking the reference style example, our models achieve higher or comparable \texttt{cos-sim} and style opinion scores in Table~\ref{table: vctk}-\ref{table: speech qualitative}
than the oracle, which is a random sample in the dataset from the same speaker. 
These results indicate the effectiveness of mimicking the style of the given reference example over utilizing speaker embedding, which learns a generic voice of a speaker.

\subsection{Online-Handwriting Synthesis}

Online-handwriting synthesis aims to generate sequences of pen movements on a writing surface.
A handwriting sample is represented as a sequence of $(x, y, p)$ triplets, where $x$ and $y$ are the coordinates of the pen on the surface, and $p$ is a binary variable indicating whether the pen touches the surface over time. 
We apply the proposed method to a subset of a proprietary dataset collected for research and development.  
The subset consists of 600k online handwriting samples that were written by 1,500 people in English, French, German, Italian, Spanish, and Portuguese.

 \begin{figure*}[t]
	\centering
	\includegraphics[width=\linewidth]{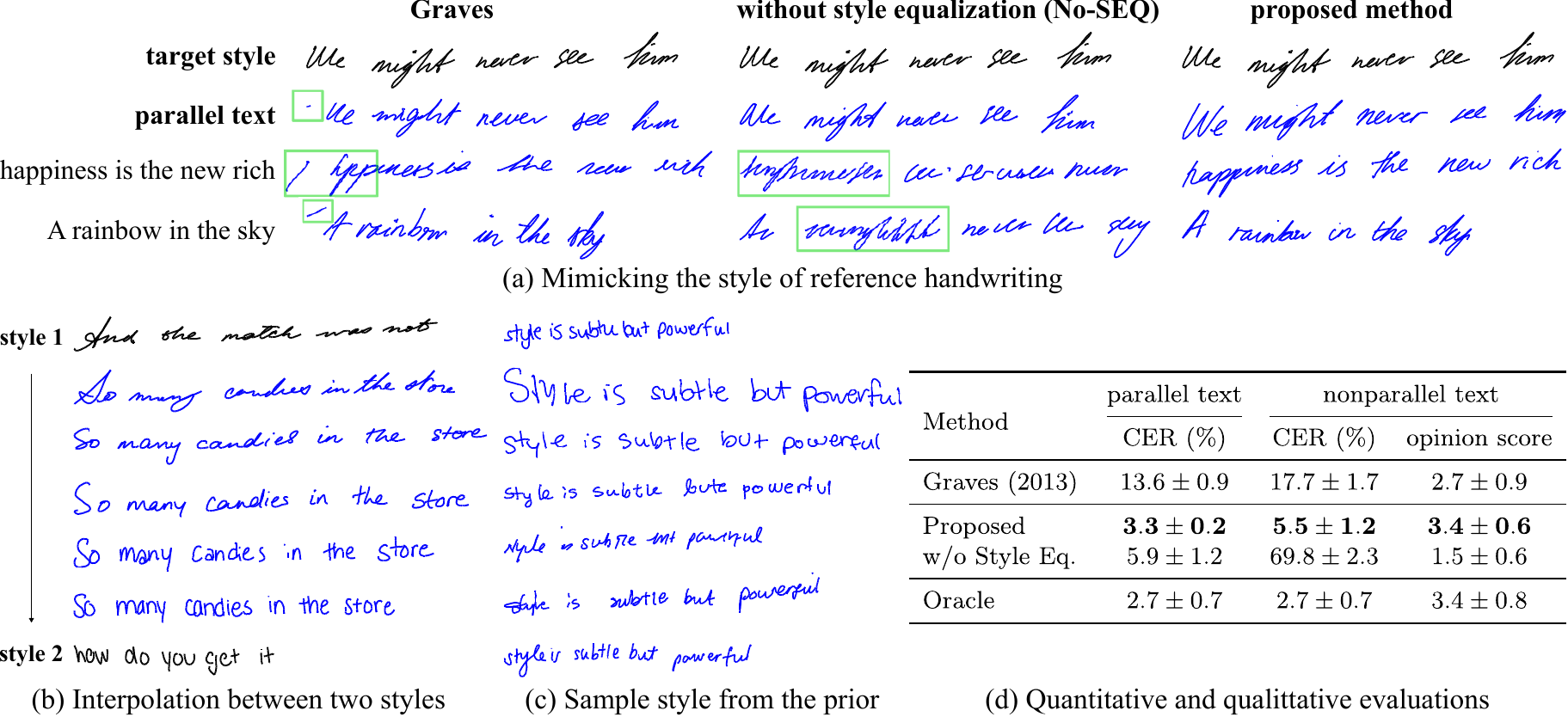}
	\vspace{-4mm}
	\caption{\textbf{Handwriting generation results and evaluation.} The reference style examples are shown in black, and the outputs are shown in blue. %
	}
	\label{figure: handwriting comparison}
	\label{figure: handwriting interpolation}
\end{figure*}

\paragraph{Baselines and metrics.}

While there exist many handwriting synthesizers, most of them requires writer labels~\citep{kotani2020generating,bhunia2021handwriting,kang2020ganwriting} or character segmentation~\citep{davis2020text,aksan2018deepwriting,kotani2020generating} (Table~\ref{table: related work}). 
We compare with the method proposed by \citet{graves2013generating} that uses priming for style encoding; since the proposed style equalization is a training mechanism, we also compare with an ablation of our model trained \textit{without} style equalization.
Note that all these models use the same decoder and content attention.  
Similar to speech, we measure content generation error as Character Error rate (CER) with a pretrained handwriting recognizer, and we conduct style-opinion-score study on 12 users and collected 320 responses.
We also provide an oracle where we select a different real handwriting sample from the same writer from the dataset and compute the metrics.

\paragraph{Results.}

\figref{figure: handwriting comparison} shows synthesized handwriting from each of the methods on unseen style examples.
As can be seen, while \citet{graves2013generating} with priming is an effective method to replicate the reference style, it often outputs artifacts at the beginning due to the residual information in the hidden state.  %
The model without style equalization produces high-quality replication in the parallel setting; however, it suffers severely under the nonparallel setting and produces wrong content.
By training with the proposed style equalization, the CER in the nonparallel setting improved significantly.
We demonstrate style interpolate in \figref{figure: handwriting interpolation}b and results with styles sampled from the learned prior distribution in \figref{figure: handwriting interpolation}c.

Note that due to privacy reasons, the handwriting reference examples shown in the paper and the supplemental website are synthetic.
They are close reproductions of unseen real styles using a generative model with a different architecture.
The generations shown here are very similar when real samples are used as style input. 
All evaluations reported in \figref{figure: handwriting interpolation}d use real unseen style examples.

\begin{figure}[t]
	\centering
	\vspace{1.5mm}
	\begin{subfigure}[t]{0.48\linewidth}
		\centering
		\includegraphics[width=\linewidth]{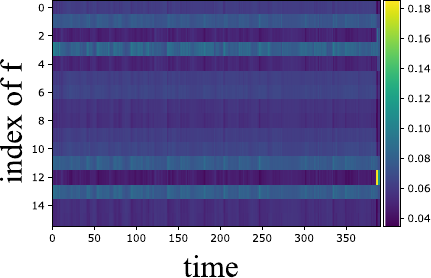}
		\caption{\scriptsize 100\%, $\c^r = \c$}
	\end{subfigure}
	~
	\begin{subfigure}[t]{0.48\linewidth}
		\centering
		\includegraphics[width=\linewidth]{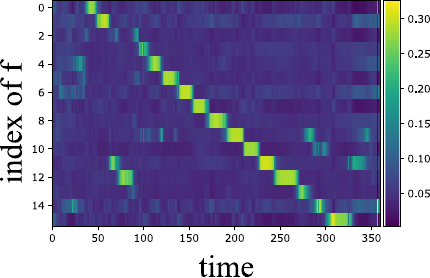}
		\caption{\scriptsize 50\%, $\c^r = \c$}
	\end{subfigure}
	\\
	\begin{subfigure}[t]{0.48\linewidth}
		\centering
		\includegraphics[width=\linewidth]{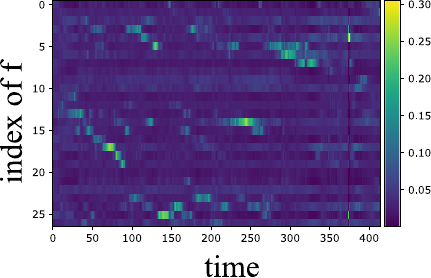}
		\caption{\scriptsize 50\%, $\c^r \neq \c$}
	\end{subfigure}
	~
	\begin{subfigure}[t]{0.48\linewidth}
		\centering
		\includegraphics[width=\linewidth]{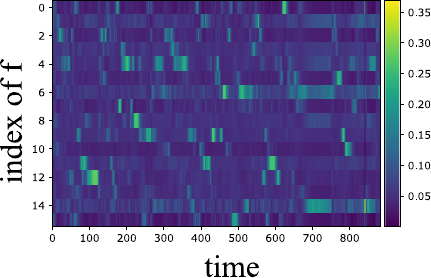}
		\caption{\scriptsize 50\%, $\c^r {=} \c$, $\x^r$ transformed}
	\end{subfigure}
	\vspace{-1.5mm}
	\caption{Style attention weights of two models trained on LibriTTS during inference on unseen styles. (a) style equalization always applied during training (100\%) and (b,c,d) style equalization applied to 50\% of the batches. All share the same input content $\c$, (a,b,d) share the same input style, but (d) transforms the input style to that of (c).  }
	\label{figure: style attention}
\end{figure}

\subsection{Analysis of Style Attention}
\label{sec: style bottleneck}

Finally, we analyze how our model utilizes the time-dependent style information contained in the reference example $\x^r$ by examining the attention weights of the style attention module.
As discussed in \secref{sec: style equalization}, while our style representation is a time-dependent sequence, style equalization can only transfer time-independent global style information from the style reference. 
Therefore, when we apply style equalization to all the samples during training, we learn a time-independent representation. 
This can be seen from \figref{figure: style attention}a where the attention weights become constant over time during inference.

We encourage our style encoder to learn the time-dependent aspects of style by applying style equalization only to 50\% of the training samples.
As can be seen from the time-varying attention weights in \figref{figure: style attention}b-d, this training procedure enables the style encoder to utilize style information more efficiently by focusing time instances with similar signal and context. 
For example, when the style reference contains the target content, \ie, $\c^r = \c$ (\figref{figure: style attention}b), the attention weights form a block-diagonal pattern, indicating the model focuses to the time instances of the style signal that match the current content and context. 
When $\c^r \neq \c$ (\figref{figure: style attention}c), the weights are still well localized over time to gather pieces of time-dependent style information from matching signal in $\x^r$. 
Last, we transform the parallel style (\figref{figure: style attention}b) to that of a nonparallel one  (\figref{figure: style attention}c) during inference in \figref{figure: style attention}d.  
Comparing \figref{figure: style attention}b \& d, the diagonal shape disappears, indicating the style transform reduces the source content information. 

The flexibility to apply/remove style equalization (and hence the representation bottleneck) during training is one of the main differences compared to \citep{wang2018style,hsu2018hierarchical}, which always apply the style bottleneck.

\section{Conclusion}

This paper proposes a simple but effective training strategy, style equalization, to mitigate the training-inference mismatch and learn a generative sequence model where style and content can be controlled separately without utilizing any style supervision. 
We demonstrate 1) replication of styles from a single style reference, 2) interpolation between two reference styles, and 3) generating new styles. 
Experiments on both speech and handwriting domains show the effectiveness of the proposed method in mitigating the training-inference mismatch, which enables our models to obtain high-quality synthesis results.

\bibliography{main}

\begin{thebibliography}{83}
\providecommand{\natexlab}[1]{#1}
\providecommand{\url}[1]{\texttt{#1}}
\expandafter\ifx\csname urlstyle\endcsname\relax
  \providecommand{\doi}[1]{doi: #1}\else
  \providecommand{\doi}{doi: \begingroup \urlstyle{rm}\Url}\fi

\bibitem[Aksan et~al.(2018)Aksan, Pece, and Hilliges]{aksan2018deepwriting}
Aksan, E., Pece, F., and Hilliges, O.
\newblock Deepwriting: Making digital ink editable via deep generative
  modeling.
\newblock In \emph{Conference on Human Factors in Computing Systems {(CHI)}},
  pp.\  1--14, 2018.

\bibitem[Akuzawa et~al.(2018)Akuzawa, Iwasawa, and
  Matsuo]{akuzawa2018expressive}
Akuzawa, K., Iwasawa, Y., and Matsuo, Y.
\newblock Expressive speech synthesis via modeling expressions with variational
  autoencoder.
\newblock In \emph{Interspeech}, 2018.

\bibitem[Alegre et~al.(2013)Alegre, Amehraye, and Evans]{alegre2013spoofing}
Alegre, F., Amehraye, A., and Evans, N.
\newblock Spoofing countermeasures to protect automatic speaker verification
  from voice conversion.
\newblock In \emph{2013 IEEE International Conference on Acoustics, Speech and
  Signal Processing}, pp.\  3068--3072. IEEE, 2013.

\bibitem[Bhunia et~al.(2021)Bhunia, Khan, Cholakkal, Anwer, Khan, and
  Shah]{bhunia2021handwriting}
Bhunia, A.~K., Khan, S., Cholakkal, H., Anwer, R.~M., Khan, F.~S., and Shah, M.
\newblock Handwriting transformers.
\newblock \emph{arXiv preprint arXiv:2104.03964}, 2021.

\bibitem[Brock et~al.(2019)Brock, Donahue, and Simonyan]{brock2018large}
Brock, A., Donahue, J., and Simonyan, K.
\newblock Large scale {GAN} training for high fidelity natural image synthesis.
\newblock In \emph{International Conference on Learning Representations
  {(ICLR)}}, 2019.

\bibitem[Brundage et~al.(2018)Brundage, Avin, Clark, Toner, Eckersley,
  Garfinkel, Dafoe, Scharre, Zeitzoff, Filar, et~al.]{brundage2018malicious}
Brundage, M., Avin, S., Clark, J., Toner, H., Eckersley, P., Garfinkel, B.,
  Dafoe, A., Scharre, P., Zeitzoff, T., Filar, B., et~al.
\newblock The malicious use of artificial intelligence: Forecasting,
  prevention, and mitigation.
\newblock \emph{arXiv preprint arXiv:1802.07228}, 2018.

\bibitem[Chan et~al.(2021)Chan, Monteiro, Kellnhofer, Wu, and
  Wetzstein]{chan2021pi}
Chan, E.~R., Monteiro, M., Kellnhofer, P., Wu, J., and Wetzstein, G.
\newblock pi-{GAN}: Periodic implicit generative adversarial networks for
  {3D}-aware image synthesis.
\newblock In \emph{{IEEE} Conference on Computer Vision and Pattern Recognition
  {(CVPR)}}, pp.\  5799--5809, 2021.

\bibitem[Chen et~al.(2021)Chen, Tan, Li, Liu, Qin, sheng zhao, and
  Liu]{chen2021adaspeech}
Chen, M., Tan, X., Li, B., Liu, Y., Qin, T., sheng zhao, and Liu, T.-Y.
\newblock {AdaSpeech}: Adaptive text to speech for custom voice.
\newblock In \emph{International Conference on Learning Representations
  {(ICLR)}}, 2021.

\bibitem[Chen et~al.(2020)Chen, Kumar, Nagarsheth, Sivaraman, and
  Khoury]{chen2020generalization}
Chen, T., Kumar, A., Nagarsheth, P., Sivaraman, G., and Khoury, E.
\newblock Generalization of audio deepfake detection.
\newblock In \emph{Proc. Odyssey 2020 The Speaker and Language Recognition
  Workshop}, pp.\  132--137, 2020.

\bibitem[Chung et~al.(2015)Chung, Kastner, Dinh, Goel, Courville, and
  Bengio]{chung2015recurrent}
Chung, J., Kastner, K., Dinh, L., Goel, K., Courville, A.~C., and Bengio, Y.
\newblock A recurrent latent variable model for sequential data.
\newblock \emph{Advances in Neural Information Processing Systems {(NeurIPS)}},
  2015.

\bibitem[Chung et~al.(2018)Chung, Nagrani, and Zisserman]{chung2018voxceleb2}
Chung, J.~S., Nagrani, A., and Zisserman, A.
\newblock {VoxCeleb2}: Deep speaker recognition.
\newblock \emph{Proc. Interspeech}, pp.\  1086--1090, 2018.

\bibitem[Collins et~al.(2020)Collins, Bala, Price, and
  Susstrunk]{Collins_2020_CVPR}
Collins, E., Bala, R., Price, B., and Susstrunk, S.
\newblock Editing in style: Uncovering the local semantics of {GAN}s.
\newblock In \emph{{IEEE} Conference on Computer Vision and Pattern Recognition
  {(CVPR)}}, 2020.

\bibitem[Davis et~al.(2020)Davis, Tensmeyer, Price, Wigington, Morse, and
  Jain]{davis2020text}
Davis, B., Tensmeyer, C., Price, B., Wigington, C., Morse, B., and Jain, R.
\newblock Text and style conditioned {GAN} for generation of offline
  handwriting lines.
\newblock In \emph{British Machine Vision Conference {(BMVC)}}, 2020.

\bibitem[Deng et~al.(2019)Deng, Guo, Xue, and Zafeiriou]{deng2019arcface}
Deng, J., Guo, J., Xue, N., and Zafeiriou, S.
\newblock Arcface: Additive angular margin loss for deep face recognition.
\newblock In \emph{{IEEE} Conference on Computer Vision and Pattern Recognition
  {(CVPR)}}, 2019.

\bibitem[Dhariwal et~al.(2020)Dhariwal, Jun, Payne, Kim, Radford, and
  Sutskever]{dhariwal2020jukebox}
Dhariwal, P., Jun, H., Payne, C., Kim, J.~W., Radford, A., and Sutskever, I.
\newblock Jukebox: A generative model for music.
\newblock \emph{arXiv preprint arXiv:2005.00341}, 2020.

\bibitem[Dolhansky et~al.(2020)Dolhansky, Bitton, Pflaum, Lu, Howes, Wang, and
  Canton~Ferrer]{dolhansky2020deepfake}
Dolhansky, B., Bitton, J., Pflaum, B., Lu, J., Howes, R., Wang, M., and
  Canton~Ferrer, C.
\newblock The deepfake detection challenge dataset.
\newblock \emph{arXiv e-prints}, pp.\  arXiv--2006, 2020.

\bibitem[Donahue et~al.(2020)Donahue, Dieleman, Bi{\'n}kowski, Elsen, and
  Simonyan]{donahue2020end}
Donahue, J., Dieleman, S., Bi{\'n}kowski, M., Elsen, E., and Simonyan, K.
\newblock End-to-end adversarial text-to-speech.
\newblock \emph{arXiv preprint arXiv:2006.03575}, 2020.

\bibitem[Esser et~al.(2019)Esser, Haux, and Ommer]{esser2019unsupervised}
Esser, P., Haux, J., and Ommer, B.
\newblock Unsupervised robust disentangling of latent characteristics for image
  synthesis.
\newblock In \emph{{IEEE} International Conference on Computer Vision
  {(ICCV)}}, 2019.

\bibitem[Esser et~al.(2020)Esser, Rombach, and Ommer]{Esser_2020_CVPR}
Esser, P., Rombach, R., and Ommer, B.
\newblock A disentangling invertible interpretation network for explaining
  latent representations.
\newblock In \emph{{IEEE} Conference on Computer Vision and Pattern Recognition
  {(CVPR)}}, June 2020.

\bibitem[Evans et~al.(2013)Evans, Kinnunen, and Yamagishi]{evans2013spoofing}
Evans, N.~W., Kinnunen, T., and Yamagishi, J.
\newblock Spoofing and countermeasures for automatic speaker verification.
\newblock In \emph{Interspeech}, pp.\  925--929, 2013.

\bibitem[Gatys et~al.(2016)Gatys, Ecker, and Bethge]{gatys2016neural}
Gatys, L., Ecker, A., and Bethge, M.
\newblock A neural algorithm of artistic style.
\newblock \emph{Journal of Vision}, 16\penalty0 (12):\penalty0 326--326, 2016.

\bibitem[Gibiansky et~al.(2017)Gibiansky, Arik, Diamos, Miller, Peng, Ping,
  Raiman, and Zhou]{arik2017deep}
Gibiansky, A., Arik, S., Diamos, G., Miller, J., Peng, K., Ping, W., Raiman,
  J., and Zhou, Y.
\newblock Deep voice 2: Multi-speaker neural text-to-speech.
\newblock In \emph{Advances in Neural Information Processing Systems
  {(NeurIPS)}}, pp.\  2966–2974, 2017.

\bibitem[Goetschalckx et~al.(2019)Goetschalckx, Andonian, Oliva, and
  Isola]{GANalyze_GoetschalckxAOI19}
Goetschalckx, L., Andonian, A., Oliva, A., and Isola, P.
\newblock {GAN}alyze: Toward visual definitions of cognitive image properties.
\newblock In \emph{{IEEE} International Conference on Computer Vision
  {(ICCV)}}, 2019.

\bibitem[Graves(2013)]{graves2013generating}
Graves, A.
\newblock Generating sequences with recurrent neural networks.
\newblock \emph{arXiv preprint arXiv:1308.0850}, 2013.

\bibitem[G{\"u}era \& Delp(2018)G{\"u}era and Delp]{guera2018deepfake}
G{\"u}era, D. and Delp, E.~J.
\newblock Deepfake video detection using recurrent neural networks.
\newblock In \emph{2018 15th IEEE international conference on advanced video
  and signal based surveillance (AVSS)}, pp.\  1--6. IEEE, 2018.

\bibitem[H\"{a}rk\"{o}nen et~al.(2020)H\"{a}rk\"{o}nen, Hertzmann, Lehtinen,
  and Paris]{HarkonenGANSpace2020}
H\"{a}rk\"{o}nen, E., Hertzmann, A., Lehtinen, J., and Paris, S.
\newblock {GANSpace}: Discovering interpretable {GAN} controls.
\newblock In \emph{Advances in Neural Information Processing Systems
  {(NeurIPS)}}, 2020.

\bibitem[Hayashi et~al.(2020)Hayashi, Yamamoto, Inoue, Yoshimura, Watanabe,
  Toda, Takeda, Zhang, and Tan]{hayashi2020espnet}
Hayashi, T., Yamamoto, R., Inoue, K., Yoshimura, T., Watanabe, S., Toda, T.,
  Takeda, K., Zhang, Y., and Tan, X.
\newblock {Espnet-{TTS}}: Unified, reproducible, and integratable open source
  end-to-end text-to-speech toolkit.
\newblock In \emph{{IEEE} International Conference on Acoustics, Speech and
  Signal Processing{(ICASSP)}}, 2020.

\bibitem[Hendrycks \& Gimpel(2016)Hendrycks and Gimpel]{hendrycks2016gaussian}
Hendrycks, D. and Gimpel, K.
\newblock Gaussian error linear units ({gelus}).
\newblock \emph{arXiv preprint arXiv:1606.08415}, 2016.

\bibitem[Henter et~al.(2018)Henter, Lorenzo-Trueba, Wang, and
  Yamagishi]{henter2018deep}
Henter, G.~E., Lorenzo-Trueba, J., Wang, X., and Yamagishi, J.
\newblock Deep encoder-decoder models for unsupervised learning of controllable
  speech synthesis.
\newblock \emph{arXiv preprint arXiv:1807.11470}, 2018.

\bibitem[Hsu et~al.(2018)Hsu, Zhang, Weiss, Zen, Wu, Wang, Cao, Jia, Chen,
  Shen, et~al.]{hsu2018hierarchical}
Hsu, W.-N., Zhang, Y., Weiss, R.~J., Zen, H., Wu, Y., Wang, Y., Cao, Y., Jia,
  Y., Chen, Z., Shen, J., et~al.
\newblock Hierarchical generative modeling for controllable speech synthesis.
\newblock In \emph{International Conference on Learning Representations
  {(ICLR)}}, 2018.

\bibitem[Hu et~al.(2020)Hu, Shrivastava, Tuzel, and Dhir]{hu2020unsupervised}
Hu, T.-Y., Shrivastava, A., Tuzel, O., and Dhir, C.
\newblock Unsupervised style and content separation by minimizing mutual
  information for speech synthesis.
\newblock In \emph{{IEEE} International Conference on Acoustics, Speech and
  Signal Processing{(ICASSP)}}, 2020.

\bibitem[Hua et~al.(2016)Hua, Huang, Shi, Goh, and Thing]{hua2016twenty}
Hua, G., Huang, J., Shi, Y.~Q., Goh, J., and Thing, V.~L.
\newblock Twenty years of digital audio watermarking --- a comprehensive
  review.
\newblock \emph{Signal processing}, 128:\penalty0 222--242, 2016.

\bibitem[Hutchinson(1989)]{hutchinson1989stochastic}
Hutchinson, M.~F.
\newblock A stochastic estimator of the trace of the influence matrix for
  laplacian smoothing splines.
\newblock \emph{Communications in Statistics-Simulation and Computation},
  18\penalty0 (3):\penalty0 1059--1076, 1989.

\bibitem[Islam et~al.(2019)Islam, Jia, and Bruce]{islam2019much}
Islam, M.~A., Jia, S., and Bruce, N.~D.
\newblock How much position information do convolutional neural networks
  encode?
\newblock In \emph{International Conference on Learning Representations
  {(ICLR)}}, 2019.

\bibitem[Isola et~al.(2017)Isola, Zhu, Zhou, and Efros]{isola2017image}
Isola, P., Zhu, J.-Y., Zhou, T., and Efros, A.~A.
\newblock Image-to-image translation with conditional adversarial networks.
\newblock In \emph{{IEEE} Conference on Computer Vision and Pattern Recognition
  {(CVPR)}}, pp.\  1125--1134, 2017.

\bibitem[Jia et~al.(2018)Jia, Zhang, Weiss, Wang, Shen, Ren, Chen, Nguyen,
  Pang, Moreno, et~al.]{jia2018transfer}
Jia, Y., Zhang, Y., Weiss, R.~J., Wang, Q., Shen, J., Ren, F., Chen, Z.,
  Nguyen, P., Pang, R., Moreno, I.~L., et~al.
\newblock Transfer learning from speaker verification to multispeaker
  text-to-speech synthesis.
\newblock In \emph{Advances in Neural Information Processing Systems
  {(NeurIPS)}}, pp.\  4485–4495, 2018.

\bibitem[Kameoka et~al.(2018)Kameoka, Kaneko, Tanaka, and
  Hojo]{kameoka2018stargan}
Kameoka, H., Kaneko, T., Tanaka, K., and Hojo, N.
\newblock Star{GAN}-{VC}: Non-parallel many-to-many voice conversion using star
  generative adversarial networks.
\newblock In \emph{Spoken Language Technology Workshop (SLT)}, pp.\  266--273.
  IEEE, 2018.

\bibitem[Kameoka et~al.(2020)Kameoka, Huang, Tanaka, Kaneko, Hojo, and
  Toda]{kameoka2020many}
Kameoka, H., Huang, W.-C., Tanaka, K., Kaneko, T., Hojo, N., and Toda, T.
\newblock Many-to-many voice transformer network.
\newblock \emph{IEEE/ACM Transactions on Audio, Speech, and Language
  Processing}, 2020.

\bibitem[kamo naoyuki(2021)]{kamo_naoyuki_2021_4604066}
kamo naoyuki.
\newblock {ESPnet2 pretrained model,
  \url{kamo-naoyuki/librispeech\_asr\_train\_asr\_conformer6\_n\_fft512\_hop\_length256\_r
  aw\_en\_bpe5000\_scheduler\_confwarmup\_steps40000\_opti
  m\_conflr0.0025\_sp\_valid.acc.ave, fs=16k, lang=en}}, March 2021.
\newblock URL \url{https://doi.org/10.5281/zenodo.4604066}.

\bibitem[Kaneko \& Kameoka(2018)Kaneko and Kameoka]{kaneko2018cyclegan}
Kaneko, T. and Kameoka, H.
\newblock Cycle{GAN}-{VC}: Non-parallel voice conversion using cycle-consistent
  adversarial networks.
\newblock In \emph{European Signal Processing Conference (EUSIPCO)}, pp.\
  2100--2104. IEEE, 2018.

\bibitem[Kaneko et~al.(2019{\natexlab{a}})Kaneko, Kameoka, Tanaka, and
  Hojo]{kaneko2019cyclegan}
Kaneko, T., Kameoka, H., Tanaka, K., and Hojo, N.
\newblock Cycle{GAN}-{VC}2: Improved cycle{GAN}-based non-parallel voice
  conversion.
\newblock In \emph{{IEEE} International Conference on Acoustics, Speech and
  Signal Processing{(ICASSP)}}, pp.\  6820--6824. IEEE, 2019{\natexlab{a}}.

\bibitem[Kaneko et~al.(2019{\natexlab{b}})Kaneko, Kameoka, Tanaka, and
  Hojo]{kaneko2019stargan}
Kaneko, T., Kameoka, H., Tanaka, K., and Hojo, N.
\newblock Star{GAN}-{VC}2: Rethinking conditional methods for star{GAN}-based
  voice conversion.
\newblock In \emph{{INTERSPEECH}}, 2019{\natexlab{b}}.

\bibitem[Kang et~al.(2020)Kang, Riba, Wang, Rusi{\~n}ol, Forn{\'e}s, and
  Villegas]{kang2020ganwriting}
Kang, L., Riba, P., Wang, Y., Rusi{\~n}ol, M., Forn{\'e}s, A., and Villegas, M.
\newblock Ganwriting: Content-conditioned generation of styled handwritten word
  images.
\newblock In \emph{European Conference on Computer Vision {(ECCV)}}, pp.\
  273--289. Springer, 2020.

\bibitem[Karras et~al.(2020)Karras, Laine, Aittala, Hellsten, Lehtinen, and
  Aila]{Karras_stylegan_v2_2020}
Karras, T., Laine, S., Aittala, M., Hellsten, J., Lehtinen, J., and Aila, T.
\newblock Analyzing and improving the image quality of {StyleGAN}.
\newblock In \emph{{IEEE} Conference on Computer Vision and Pattern Recognition
  {(CVPR)}}, 2020.

\bibitem[Karras et~al.(2021)Karras, Aittala, Laine, H{\"a}rk{\"o}nen, Hellsten,
  Lehtinen, and Aila]{karras2021alias}
Karras, T., Aittala, M., Laine, S., H{\"a}rk{\"o}nen, E., Hellsten, J.,
  Lehtinen, J., and Aila, T.
\newblock Alias-free generative adversarial networks.
\newblock \emph{Advances in Neural Information Processing Systems {(NeurIPS)}},
  34, 2021.

\bibitem[Kazemi et~al.(2019)Kazemi, Iranmanesh, and Nasrabadi]{kazemi2019style}
Kazemi, H., Iranmanesh, S.~M., and Nasrabadi, N.
\newblock Style and content disentanglement in generative adversarial networks.
\newblock In \emph{IEEE Winter Conference on Applications of Computer Vision
  (WACV)}, pp.\  848--856, 2019.

\bibitem[Kim et~al.(2020)Kim, Kim, Kong, and Yoon]{kim2020glow}
Kim, J., Kim, S., Kong, J., and Yoon, S.
\newblock Glow-{TTS}: A generative flow for text-to-speech via monotonic
  alignment search.
\newblock \emph{Advances in Neural Information Processing Systems {(NeurIPS)}},
  33, 2020.

\bibitem[Kingma \& Ba(2015)Kingma and Ba]{kingma2014adam}
Kingma, D.~P. and Ba, J.
\newblock Adam: A method for stochastic optimization.
\newblock In \emph{International Conference on Learning Representations
  {(ICLR)}}, 2015.
\newblock URL \url{http://arxiv.org/abs/1412.6980}.

\bibitem[Kingma \& Welling(2014)Kingma and Welling]{KingmaW13VAE}
Kingma, D.~P. and Welling, M.
\newblock Auto-encoding variational bayes.
\newblock In \emph{International Conference on Learning Representations
  {(ICLR)}}, 2014.

\bibitem[Kinnunen et~al.(2020)Kinnunen, Delgado, Evans, Lee, Vestman, Nautsch,
  Todisco, Wang, Sahidullah, Yamagishi, et~al.]{kinnunen2020tandem}
Kinnunen, T., Delgado, H., Evans, N., Lee, K.~A., Vestman, V., Nautsch, A.,
  Todisco, M., Wang, X., Sahidullah, M., Yamagishi, J., et~al.
\newblock Tandem assessment of spoofing countermeasures and automatic speaker
  verification: Fundamentals.
\newblock \emph{IEEE/ACM Transactions on Audio, Speech, and Language
  Processing}, 28:\penalty0 2195--2210, 2020.

\bibitem[Kotani et~al.(2020)Kotani, Tellex, and Tompkin]{kotani2020generating}
Kotani, A., Tellex, S., and Tompkin, J.
\newblock Generating handwriting via decoupled style descriptors.
\newblock In \emph{European Conference on Computer Vision {(ECCV)}}, pp.\
  764--780. Springer, 2020.

\bibitem[Kotovenko et~al.(2019)Kotovenko, Sanakoyeu, Lang, and
  Ommer]{kotovenko2019content}
Kotovenko, D., Sanakoyeu, A., Lang, S., and Ommer, B.
\newblock Content and style disentanglement for artistic style transfer.
\newblock In \emph{{IEEE} International Conference on Computer Vision
  {(ICCV)}}, pp.\  4422--4431, 2019.

\bibitem[Kwon \& Ye(2021)Kwon and Ye]{kwon2021diagonal}
Kwon, G. and Ye, J.~C.
\newblock Diagonal attention and style-based gan for content-style
  disentanglement in image generation and translation.
\newblock In \emph{{IEEE} International Conference on Computer Vision
  {(ICCV)}}, pp.\  13980--13989, 2021.

\bibitem[Lample et~al.(2017)Lample, Zeghidour, Usunier, Bordes, DENOYER, and
  Ranzato]{NIPS2017_faderNet}
Lample, G., Zeghidour, N., Usunier, N., Bordes, A., DENOYER, L., and Ranzato,
  M.~A.
\newblock Fader networks:manipulating images by sliding attributes.
\newblock In \emph{Advances in Neural Information Processing Systems
  {(NeurIPS)}}, 2017.

\bibitem[Lee et~al.(2021)Lee, Kim, Chung, and Lee]{lee2021voicemixer}
Lee, S.-H., Kim, J.-H., Chung, H., and Lee, S.-W.
\newblock Voicemixer: Adversarial voice style mixup.
\newblock \emph{Advances in Neural Information Processing Systems {(NeurIPS)}},
  34, 2021.

\bibitem[Lyu(2020)]{lyu2020deepfake}
Lyu, S.
\newblock Deepfake detection: Current challenges and next steps.
\newblock In \emph{2020 IEEE International Conference on Multimedia \& Expo
  Workshops (ICMEW)}, pp.\  1--6. IEEE, 2020.

\bibitem[Ma et~al.(2018)Ma, Mcduff, and Song]{ma2018neural}
Ma, S., Mcduff, D., and Song, Y.
\newblock Neural {TTS} stylization with adversarial and collaborative games.
\newblock In \emph{International Conference on Learning Representations
  {(ICLR)}}, 2018.

\bibitem[McAuliffe et~al.(2017)McAuliffe, Socolof, Mihuc, Wagner, and
  Sonderegger]{mcauliffe2017montreal}
McAuliffe, M., Socolof, M., Mihuc, S., Wagner, M., and Sonderegger, M.
\newblock Montreal forced aligner: Trainable text-speech alignment using kaldi.
\newblock In \emph{Interspeech}, volume 2017, pp.\  498--502, 2017.

\bibitem[Meng et~al.(2021)Meng, Song, Song, Zhao, and Ermon]{meng2021improved}
Meng, C., Song, J., Song, Y., Zhao, S., and Ermon, S.
\newblock Improved autoregressive modeling with distribution smoothing.
\newblock In \emph{International Conference on Learning Representations
  {(ICLR)}}, 2021.

\bibitem[Pavllo et~al.(2020)Pavllo, Lucchi, and Hofmann]{pavllo2020controlling}
Pavllo, D., Lucchi, A., and Hofmann, T.
\newblock Controlling style and semantics in weakly-supervised image
  generation.
\newblock In \emph{European Conference on Computer Vision {(ECCV)}}, pp.\
  482--499. Springer, 2020.

\bibitem[Ping et~al.(2018)Ping, Peng, Gibiansky, Arik, Kannan, Narang, Raiman,
  and Miller]{ping2018deep}
Ping, W., Peng, K., Gibiansky, A., Arik, S.~{\"O}., Kannan, A., Narang, S.,
  Raiman, J., and Miller, J.
\newblock Deep voice 3: Scaling text-to-speech with convolutional sequence
  learning.
\newblock In \emph{International Conference on Learning Representations
  {(ICLR)}}, 2018.

\bibitem[Prenger et~al.(2019)Prenger, Valle, and
  Catanzaro]{prenger2019waveglow}
Prenger, R., Valle, R., and Catanzaro, B.
\newblock Waveglow: A flow-based generative network for speech synthesis.
\newblock In \emph{{IEEE} International Conference on Acoustics, Speech and
  Signal Processing{(ICASSP)}}, pp.\  3617--3621. IEEE, 2019.

\bibitem[Qian et~al.(2019)Qian, Zhang, Chang, Yang, and
  Hasegawa-Johnson]{qian2019autovc}
Qian, K., Zhang, Y., Chang, S., Yang, X., and Hasegawa-Johnson, M.
\newblock Auto{VC}: Zero-shot voice style transfer with only autoencoder loss.
\newblock In \emph{International Conference on Machine Learning {(ICML)}}, pp.\
   5210--5219. PMLR, 2019.

\bibitem[Qian et~al.(2020)Qian, Zhang, Chang, Hasegawa-Johnson, and
  Cox]{qian2020unsupervised}
Qian, K., Zhang, Y., Chang, S., Hasegawa-Johnson, M., and Cox, D.
\newblock Unsupervised speech decomposition via triple information bottleneck.
\newblock In \emph{International Conference on Machine Learning {(ICML)}}, pp.\
   7836--7846. PMLR, 2020.

\bibitem[Ramachandran et~al.(2018)Ramachandran, Zoph, and
  Le]{ramachandran2017searching}
Ramachandran, P., Zoph, B., and Le, Q.
\newblock Searching for activation functions.
\newblock In \emph{International Conference on Learning Representations
  {(ICLR)}}, 2018.

\bibitem[Ren et~al.(2020)Ren, Hu, Tan, Qin, Zhao, Zhao, and
  Liu]{ren2020fastspeech}
Ren, Y., Hu, C., Tan, X., Qin, T., Zhao, S., Zhao, Z., and Liu, T.-Y.
\newblock Fastspeech 2: Fast and high-quality end-to-end text to speech.
\newblock \emph{arXiv preprint arXiv:2006.04558}, 2020.

\bibitem[Shen et~al.(2018)Shen, Pang, Weiss, Schuster, Jaitly, Yang, Chen,
  Zhang, Wang, Skerrv-Ryan, et~al.]{shen2018natural}
Shen, J., Pang, R., Weiss, R.~J., Schuster, M., Jaitly, N., Yang, Z., Chen, Z.,
  Zhang, Y., Wang, Y., Skerrv-Ryan, R., et~al.
\newblock Natural {TTS} synthesis by conditioning wave{N}et on mel spectrogram
  predictions.
\newblock In \emph{{IEEE} International Conference on Acoustics, Speech and
  Signal Processing{(ICASSP)}}, pp.\  4779--4783, 2018.

\bibitem[Shen et~al.(2020)Shen, Gu, Tang, and Zhou]{Shen_2020_CVPR}
Shen, Y., Gu, J., Tang, X., and Zhou, B.
\newblock Interpreting the latent space of {GAN}s for semantic face editing.
\newblock In \emph{{IEEE} Conference on Computer Vision and Pattern Recognition
  {(CVPR)}}, 2020.

\bibitem[Singh et~al.(2019)Singh, Ojha, and Lee]{singh_cvpr2019}
Singh, K.~K., Ojha, U., and Lee, Y.~J.
\newblock Fine{GAN}: Unsupervised hierarchical disentanglement for fine-grained
  object generation and discovery.
\newblock In \emph{{IEEE} Conference on Computer Vision and Pattern Recognition
  {(CVPR)}}, 2019.

\bibitem[Skerry-Ryan et~al.(2018)Skerry-Ryan, Battenberg, Xiao, Wang, Stanton,
  Shor, Weiss, Clark, and Saurous]{skerry2018towards}
Skerry-Ryan, R., Battenberg, E., Xiao, Y., Wang, Y., Stanton, D., Shor, J.,
  Weiss, R., Clark, R., and Saurous, R.~A.
\newblock Towards end-to-end prosody transfer for expressive speech synthesis
  with tacotron.
\newblock In \emph{International Conference on Machine Learning {(ICML)}}, pp.\
   4693--4702, 2018.

\bibitem[Snyder et~al.(2018)Snyder, Garcia-Romero, Sell, Povey, and
  Khudanpur]{snyder2018x}
Snyder, D., Garcia-Romero, D., Sell, G., Povey, D., and Khudanpur, S.
\newblock X-vectors: Robust dnn embeddings for speaker recognition.
\newblock In \emph{{IEEE} International Conference on Acoustics, Speech and
  Signal Processing{(ICASSP)}}, pp.\  5329--5333. IEEE, 2018.

\bibitem[Sun et~al.(2020{\natexlab{a}})Sun, Zhang, Weiss, Cao, Zen, Rosenberg,
  Ramabhadran, and Wu]{sun2020generating}
Sun, G., Zhang, Y., Weiss, R.~J., Cao, Y., Zen, H., Rosenberg, A., Ramabhadran,
  B., and Wu, Y.
\newblock Generating diverse and natural text-to-speech samples using a
  quantized fine-grained {VAE} and autoregressive prosody prior.
\newblock In \emph{{IEEE} International Conference on Acoustics, Speech and
  Signal Processing{(ICASSP)}}, pp.\  6699--6703, 2020{\natexlab{a}}.

\bibitem[Sun et~al.(2020{\natexlab{b}})Sun, Zhang, Weiss, Cao, Zen, and
  Wu]{sun2020fully}
Sun, G., Zhang, Y., Weiss, R.~J., Cao, Y., Zen, H., and Wu, Y.
\newblock Fully-hierarchical fine-grained prosody modeling for interpretable
  speech synthesis.
\newblock In \emph{{IEEE} International Conference on Acoustics, Speech and
  Signal Processing{(ICASSP)}}, pp.\  6264--6268. IEEE, 2020{\natexlab{b}}.

\bibitem[Valle et~al.(2020)Valle, Li, Prenger, and
  Catanzaro]{valle2020mellotron}
Valle, R., Li, J., Prenger, R., and Catanzaro, B.
\newblock Mellotron: Multispeaker expressive voice synthesis by conditioning on
  rhythm, pitch and global style tokens.
\newblock In \emph{{IEEE} International Conference on Acoustics, Speech and
  Signal Processing{(ICASSP)}}, pp.\  6189--6193. IEEE, 2020.

\bibitem[Vaswani et~al.(2017)Vaswani, Shazeer, Parmar, Uszkoreit, Jones, Gomez,
  Kaiser, and Polosukhin]{vaswani2017attention}
Vaswani, A., Shazeer, N., Parmar, N., Uszkoreit, J., Jones, L., Gomez, A.~N.,
  Kaiser, L.~u., and Polosukhin, I.
\newblock Attention is all you need.
\newblock In Guyon, I., Luxburg, U.~V., Bengio, S., Wallach, H., Fergus, R.,
  Vishwanathan, S., and Garnett, R. (eds.), \emph{Advances in Neural
  Information Processing Systems {(NeurIPS)}}, volume~30. Curran Associates,
  Inc., 2017.
\newblock URL
  \url{https://proceedings.neurips.cc/paper/2017/file/3f5ee243547dee91fbd053c1c4a845aa-Paper.pdf}.

\bibitem[Wang et~al.(2018)Wang, Stanton, Zhang, Ryan, Battenberg, Shor, Xiao,
  Jia, Ren, and Saurous]{wang2018style}
Wang, Y., Stanton, D., Zhang, Y., Ryan, R.-S., Battenberg, E., Shor, J., Xiao,
  Y., Jia, Y., Ren, F., and Saurous, R.~A.
\newblock Style tokens: Unsupervised style modeling, control and transfer in
  end-to-end speech synthesis.
\newblock In \emph{International Conference on Machine Learning {(ICML)}}, pp.\
   5180--5189. PMLR, 2018.

\bibitem[Wu et~al.(2017)Wu, Yamagishi, Kinnunen, Hanil{\c{c}}i, Sahidullah,
  Sizov, Evans, Todisco, and Delgado]{wu2017asvspoof}
Wu, Z., Yamagishi, J., Kinnunen, T., Hanil{\c{c}}i, C., Sahidullah, M., Sizov,
  A., Evans, N., Todisco, M., and Delgado, H.
\newblock Asvspoof: the automatic speaker verification spoofing and
  countermeasures challenge.
\newblock \emph{IEEE Journal of Selected Topics in Signal Processing},
  11\penalty0 (4):\penalty0 588--604, 2017.

\bibitem[Yamagishi et~al.(2019)Yamagishi, Veaux, and
  MacDonald]{yamagishi2019vctk}
Yamagishi, J., Veaux, C., and MacDonald, K.
\newblock {CSTR VCTK Corpus: English Multi-speaker Corpus for CSTR Voice
  Cloning Toolkit (version 0.92), [sound]}.
\newblock \emph{University of Edinburgh. The Centre for Speech Technology
  Research (CSTR). https://doi.org/10.7488/ds/2645.}, 2019.

\bibitem[Zen et~al.(2019)Zen, Dang, Clark, Zhang, Weiss, Jia, Chen, and
  Wu]{zen2019libritts}
Zen, H., Dang, V., Clark, R., Zhang, Y., Weiss, R.~J., Jia, Y., Chen, Z., and
  Wu, Y.
\newblock {LibriTTS}: A corpus derived from librispeech for text-to-speech.
\newblock \emph{arXiv preprint arXiv:1904.02882}, 2019.

\bibitem[Zhang(2019)]{zhang2019making}
Zhang, R.
\newblock Making convolutional networks shift-invariant again.
\newblock In \emph{International Conference on Machine Learning {(ICML)}}, pp.\
   7324--7334. PMLR, 2019.

\bibitem[Zhang et~al.(2018)Zhang, Zhang, and Cai]{zhang2018separating}
Zhang, Y., Zhang, Y., and Cai, W.
\newblock Separating style and content for generalized style transfer.
\newblock In \emph{{IEEE} Conference on Computer Vision and Pattern Recognition
  {(CVPR)}}, pp.\  8447--8455, 2018.

\bibitem[Zhao et~al.(2020)Zhao, Huang, Tian, Yamagishi, Das, Kinnunen, Ling,
  and Toda]{zhao2020voice}
Zhao, Y., Huang, W.-C., Tian, X., Yamagishi, J., Das, R.~K., Kinnunen, T.,
  Ling, Z., and Toda, T.
\newblock Voice conversion challenge 2020: Intra-lingual semi-parallel and
  cross-lingual voice conversion.
\newblock \emph{arXiv preprint arXiv:2008.12527}, 2020.

\bibitem[Zhu et~al.(2017)Zhu, Park, Isola, and Efros]{zhu2017unpaired}
Zhu, J.-Y., Park, T., Isola, P., and Efros, A.~A.
\newblock Unpaired image-to-image translation using cycle-consistent
  adversarial networks.
\newblock In \emph{{IEEE} International Conference on Computer Vision
  {(ICCV)}}, pp.\  2223--2232, 2017.

\end{thebibliography}
\bibliographystyle{icml2022}

\newpage
\appendix
\onecolumn

\section{Broader Impact}
\label{sec: broader impact}

The technology we develop in the paper, like other artificial intelligence technologies, potentially has both positive and negative impacts to our society~\citep{brundage2018malicious}.
One potential risk associated with all generative models is creating fake digital content. 
If deployed irresponsibly, speech and handwriting synthesis could facilitate deceptive interactions, including mimicking a person's voice or handwriting in automatic speared phishing attacks, gaining security access, or directing public opinions.
Examples of responsible deployments of the technology include (but not limited to):
\begin{itemize}[leftmargin=*]
	\item A system-level authentication every time the technology is used to register style or generate output.
	\item An encryption system to protect registered style. 
	\item A watermarking system (for both speech and handwriting) such that the generations can be easily identified by human or detection systems~\citep{hua2016twenty}.
\end{itemize}
Beyond system-level security measures, technologies to identify fake digital content have also been rapidly developed~\citep{chen2020generalization,lyu2020deepfake,guera2018deepfake,dolhansky2020deepfake,alegre2013spoofing,wu2017asvspoof,evans2013spoofing,kinnunen2020tandem}.
Despite the potential negative societal impacts, we believe that the technology will have a larger positive impact, such as enabling new accessibility capabilities (\eg, helping mute people speak in their voices and paralyzed people write in their handwriting) and better human-computer interaction (\eg, by improving downstream speech and handwriting recognizers).

\section{Details of the Training Procedure}
\label{sec: tricks}

Training with style equalization is straightforward --- we jointly optimize all the model parameters by maximizing the log-likelihood lower bound as described in \equref{eq: lower bound} of the main paper.
We found the following {optional} steps to be useful to improve training and the quality of the generation.

\begin{itemize}[leftmargin=*]

\item During training, we found it helpful to add a small amount of Gaussian noise to the ground-truth $x_{t{-}1}$ that is fed back to the bottom LSTM.
Intuitively, we are simulating the noise caused by sampling the output distribution during inference at training time.
We found that it makes the inference more stable.
A similar method is proposed by \citet{meng2021improved} to learn auto-regressive models. 

\item To encourage the basis $A$ that is used by $\fphi$ in \equref{eq: T and phi} to capture a wide variety of styles, we maintain $A$ as an orthonormal basis.
We normalize each column $a_i$ in $A$ to be unit norm in the architecture and minimize $|a_i^\top a_j|^2$ for all $i$ and $j \neq i$.
The minimization is conducted by minimizing the trace of $(A^\top A)^2$, which is estimated efficiently using Hutchinson's trace estimator~\citep{hutchinson1989stochastic} with 100 random samples from a standard Normal distribution.
The estimated value is used as a regularization with loss weight equal to 1.
\end{itemize}

Overall, we optimize the following objective function
\begin{multline}
	\label{eq: full objective}
	\max_{\theta, A}  \bbE_{\substack{(\x, \c) \sim \cX \\ (\x', \c') \sim \cX \\ \n \sim \mN(0, \sigma^2)}} \  \sum_{t=1}^{M} \bbE_{z_t \sim q_\theta (z_t|\fT_\theta(\x', \fphi_\theta(\x', \x), \x_\past, \c)}  
	\log p_\theta (x_t | z_t, \x_\past + \n_\past, \c) \\ 
	- \mD_{KL} \left( q_\theta(z_t | \fT_\theta(\x', \fphi_\theta(\x', \x), \x_\past, \c) \, || \, p_\theta(z_t | \x_\past, \c) \right) - \tr((A^\top A)^2),
\end{multline}
where $\theta$ represents all network parameters (except $A$).
We use the reparameterization trick that is commonly used in variational autoencoders~\citep{KingmaW13VAE} with one sample to estimate the inner expectation in \equref{eq: full objective}, and we use ADAM~\citep{kingma2014adam} with $\beta_1=0.9$, $\beta_2 = 0.98$, and a learning rate schedule used by \citet{vaswani2017attention} with a warm-up period of $4,000$ iterations to optimize the objective function.
The learning rate increases rapidly within the warm-up period to $10^{-4}$ and decreases slowly after then.

\begin{figure}[t]
	\centering
	\includegraphics[width=\linewidth]{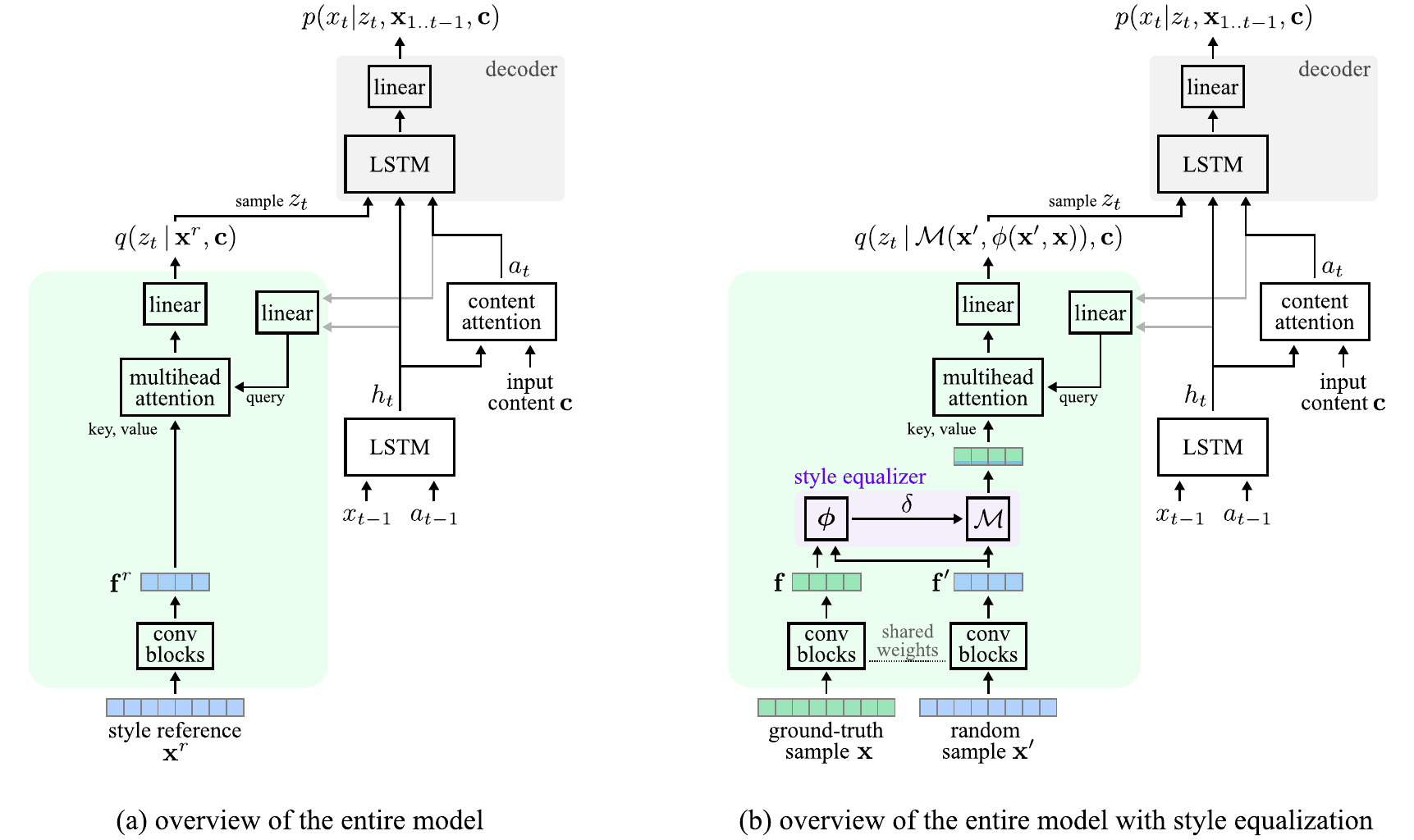}
	\caption{\textbf{Overview of the model.} 
		(a) shows an overview of the entire model without style equalization (used during inference since $\delta = \fphi(\x^r, \x^r) = 0$).  
		 It includes a style encoder (in green), a content attention, a decoder (in gray), and a LSTM at the bottom. Note that the input content $\c$ can be the output of a content-embedding network (used in speech synthesis) or one-hot encoding of characters (used in handwriting synthesis). $a_t$ is the output of the content attention at time $t$, which is a linear combination of the elements in $\c$.
		(b) shows an overview of the entire model with style equalization (used during training or during interpolation).  
		 $\fphi$ computes the vector $\delta$ that encodes the amount of style transformation between $\x'$ and $\x$. $\fT$ applies this transformation to $\x'$ to match the style of $\x$. 
		Please see Table~\ref{table: arch details} for details about individual blocks. 
 	}
	\label{figure: full arch}
\end{figure}

\renewcommand{\next}{$\rightarrow$}  
\newcommand{\blur}{blur}  %
\newcommand{\conv}[4]{conv({{#1}, {#2}, {#3}, {#4}})}  %
\newcommand{\silu}{Swish}  %
\newcommand{\dropout}[1]{dropout({#1})}  %

\begin{table}[t]
	\caption{Block formulation used in \figref{figure: full arch}}
	\label{table: arch details}
	\centering
	\setlength{\tabcolsep}{5pt}
	\begin{adjustbox}{max width=\linewidth}
		\begin{tabular}{p{23mm}l}
			\toprule
			Block name & \multicolumn{1}{c}{Architecture}   \\ 
			\midrule
			conv blocks & 
			\begin{tabular}[t]{llllllll}
				\blur & \next &  \conv{3}{$f_1$}{2}{0} & \next & \silu & \next & \dropout{$0.1$}  & \next \\
				\blur & \next &  \conv{3}{$f_2$}{2}{0} & \next & \silu & \next & \dropout{$0.1$}  & \next \\
				\blur & \next &  \conv{3}{$f_3$}{2}{0} & \next & \silu & \next & \dropout{$0.1$}  & \next \\
				\blur & \next &  \conv{3}{$f_4$}{2}{0} & \next & \silu & \next & \dropout{$0.1$}  &  \\
				\multicolumn{8}{l}{
					$
					(f_1, f_2, f_3, f_4) =
					\begin{cases}
						(32, 64, 128, 256),  &  \text{for handwriting} \\ 
						(256, 384, 512, 512), & \text{for speech} 
					\end{cases}	
					$
				} 
			\end{tabular}
			\\ \\
			multihead attention & 
			\begin{tabular}[t]{l}
				number of heads = 4 \\
				$
				\text{dimension of query, key, value} =
				\begin{cases}
					128,  &  \text{for handwriting} \\ 
					64, & \text{for VCTK}  \\
					192, & \text{for LibriTTS} 
				\end{cases}	
				$
			\end{tabular}
			\\ \\
			bottom LSTM & 
			\begin{tabular}[t]{l}
				number of layers = 1 \\
				$
				\text{dimension} =
				\begin{cases}
					512,  &  \text{for handwriting} \\ 
					2048, & \text{for speech}  
				\end{cases}	
				$
			\end{tabular}
			\\ \\
			top LSTM & 
			\begin{tabular}[t]{l}
				number of layers = 2 \\
				$
				\text{dimension} =
				\begin{cases}
					512,  &  \text{for handwriting} \\ 
					2048, & \text{for speech}  
				\end{cases}	
				$
			\end{tabular}
			\\ \\
			content attention & 
			\begin{tabular}[t]{l}
				number of Gaussian windows = 10 \\
				See \cite{graves2013generating} for the exact formulation. \\
			\end{tabular}
			\\ \\
			content encoder (speech-only) & 
			\begin{tabular}[t]{llllllll}
				\conv{5}{256}{1}{2} & \next & \silu & \next & \conv{5}{256}{1}{2} & \next & \silu & \next  \\
				\conv{5}{256}{1}{2} & \next & \silu & \next & \multicolumn{4}{l}{bidirectional LSTM (dim=256)} \\
			\end{tabular}
			\\ \\
			\midrule
			\multicolumn{2}{l}{
				\begin{tabular}[t]{l}
					\texttt{blur}: 1D low-pass filtering with kernel $[1 \ 3 \ 3 \ 1]$ \\
					\texttt{conv($k,f,s,p$)}: 1D convolution with kernel size $k$, feature dimension $f$, stride $s$, and padding $p$ \\
					\texttt{Swish}: Swish nonlinearity  \cite{ramachandran2017searching,hendrycks2016gaussian} \\
					\texttt{dropout($p$)}: dropout with probability $p$ \\
				\end{tabular}
			} 	
			\\
			\bottomrule
		\end{tabular}
	\end{adjustbox}
\end{table}

\section{Model Architecture Details}
\label{sec: model architecture details}

In this section, we provide more details about the model architecture used in the paper.
\figref{figure: full arch} shows an overview of the model, and Table \ref{table: arch details} lists the individual block formulations used in \figref{figure: full arch}.
We use the same architecture for both handwriting and speech synthesis, except for the hyper-parameters, which we list at the end of the section.

The model can be separated into a backbone and a style encoder. 
The backbone is composed of a decoder, content attention, and an LSTM at the bottom.
It is a standard model architecture and has been used and extended in many works of handwriting and speech synthesis~\citep{graves2013generating, shen2018natural, wang2018style, hsu2018hierarchical}.

The bottom LSTM is one-layer, and it accumulates information from the past outputs $\x_\past$ and the previously attended content $a_1, \dots, a_{t{-}1}$ into its hidden state $h_t$.
The content attention, which is proposed by \citet{graves2013generating}, utilizes moving Gaussian windows to calculate attention weights for the content. 
The focused content $a_t$ at time $t$ is a linear combination of the elements in $\c$ based on the attention weights.
The decoder at the top is a two-layer LSTM that takes all available information, including $h_t$, $z_t$, and $a_t$, and outputs the parameters of $p(x_t | z_t, \x_\past, \c)$.
As mentioned in \secref{sec: tricks}, we add a small Gaussian noise to $x_{t{-}1}$ that is passed to the bottom LSTM.

Our style encoder is composed of a 4-layer convolutional network and a multi-head attention layer.
Given a style reference input $\x^r$, the convolutional network extracts the feature sequence $\f^r$.
We apply low-pass filtering before every sub-sampling~\citep{zhang2019making} to avoid aliasing caused by sub-sampling in the convolutional network.
To maintain time-invariant of the convolutional network, we do not add any padding~\citep{islam2019much}. 
Given the past outputs $h_t$ and the currently focused content $c_t$, we use multi-head attention to extract relevant information from $\f^r$.  
The query vector is computed from $h_t$ and $c_t$ using a linear layer, and the key and the value vectors are the individual feature vectors contained in the sequence $\f^r$ without positional encoding. 
The intuition is that if the model plans to write a specific character or utter a specific word, it should find the information in the style reference and mimic it. 
The variational approximation $q(z_t | \cdot)$ is a multivariate Gaussian distribution with a diagonal covariance matrix.
The prior distribution $p(z_t | \x_\past, \c)$ is also modeled as a multivariate Gaussian distribution with a diagonal covariance matrix, and we a two-layer feed-forward network to compute its means and standard deviations from $h_t$ and $c_t$.
When the style equalization is used, $\fT$ and $\fphi$ are inserted into the style encoder, as shown in \figref{figure: full arch}(b).

Now we summarize the hyper-parameters  used for handwriting and speech.
Please also see Table~\ref{table: arch details}.
\begin{itemize}[leftmargin=*]
	\item For handwriting, the dimension of all LSTMs are $512$.  
	The final linear layer outputs a 122-dimensional vector, which is used to parameterize the output distribution. 
	The output distribution includes a mixture of $20$ bivariate Gaussian distributions that model the pen movement, a Bernoulli distribution for pen lifting, and a Bernoulli distribution for sequence stops.  
	The posterior and the prior Gaussian distributions are $256$-dimensional.  
	The convolutional network in the style encoder has four layers; all of them use kernel size 3, stride 2, and no padding.
	Their feature dimensions are $3 \rightarrow 32 \rightarrow 64 \rightarrow 128 \rightarrow 256$. 
	We use dropout with a dropping rate equal to 0.1 after each nonlinearity in the convolutional network.
	The multihead attention has 4 heads, the dimension of the query, key, and value vectors are all 256. 
	The dimension of $\delta$ (\ie, $k$) is $128$.  
	The input content $\c$ is represented as 200-dimensional one-hot vectors of the input characters. 
	The standard deviation of the Gaussian noise is $0.1$,  and during inference, we reduce the standard deviation of the output distribution to $0.9$ of the original one. 
	The model is trained for 100 epochs on a machine with 8 A100 GPUs, and the training took 36 hours.
	\item For speech, all LSTMs have the same $2048$ dimension. 
	The final linear layer outputs a 484-dimensional vector, which is used to parameterize the output distribution. 
	The output distribution includes a mixture of three 80-dimensional Gaussian distributions with diagonal covariance that models the mel-spectrogram and a Bernoulli distribution for sequence stops. 
	The convolutional network in the style encoder has four layers; all of them use kernel size 3, stride 2, no padding. 
	Their feature dimensions are $80 \rightarrow 256 \rightarrow 384 \rightarrow 512 \rightarrow 512$. 
	We use dropout with a dropping rate equal to 0.1 after each nonlinearity in the convolutional network.
	The multihead attention has 4 heads, the dimension of the query, key, and value vectors are all 256. 
	
	The input sentence is represented as phonemes, which contain 148 symbols.  
	We follow the pre-processing used by \citet{shen2018natural} for phoneme and mel-spectrogram extraction. 
	We also follow \citet{shen2018natural} and use a bidirectional LSTM and a convolutional network to encode the dependencies between phonemes. 
	The architecture is the same as that used by \citet{shen2018natural}.  
	The posterior and prior Gaussian distributions are $512$-dimensional, and the dimension of $\delta$ (\ie, $k$) is $64$ for VCTK dataset and $192$ for LibriTTS dataset.  
	The standard deviation of the added noise is $0.2$, and during inference, we reduce the standard deviation of the output distribution to $0.74$ of the original one. 
	The VCTK model is trained for 70 epochs on a machine with 8 A100 GPUs, and the training took 12 hours; the LibriTTS model is trained for 25 epochs on a machine with 8 A100 GPUs, and the training took 3 days.
	During inference, the LibriTTS model generates $\sim$380 mel-spectrogram frames (4.4 seconds) per second on a single A100 GPU.
\end{itemize}

\section{Style Classifier Network}
\label{sec: style classifier network}

As we mentioned in \secref{sec: speech metric}, we train a speaker classifier using the objective function proposed by \citet{deng2019arcface}.
Using the features extracted by the speaker classifier, we measure the style similarity between two waveforms using the cosine similarity between the features.
The style classifier comprises a convolutional network, an LSTM, and a linear layer that transforms the last hidden state into the feature that we use to compute the cosine similarity.
The input to the convolutional network is the 80-dimensional phoneme, which is extracted using the same procedure as the one used by \citet{shen2018natural}. 
The convolutional network has four layers; all of them use kernel size 3, stride 2, valid padding, and swish non-linearity~\citep{ramachandran2017searching}. 
Their feature dimensions are $80 \rightarrow 256 \rightarrow 384 \rightarrow 512 \rightarrow 512$.  
We use dropout with a dropping rate equal to 0.1 after each non-linearity in the convolutional network.
The LSTM has one layer, and its dimension is 512.
We split the training set of LibriTTS-all-960 into the training, validation, and test sets by a ratio of 85\%, 7.5\%, 7.5\%, respectively.
We use the same learning rate schedule and optimizer mentioned above to train the classifier.
The classifier achieves 96.5\% validation accuracy.

\section{Overview of Related Works}

\new We provide a high-level summary of various controllable sequence generative models in Table~\ref{table: related work}. 
In the table, we compare the methods on their needs of : (1) user identification or pretrained embedding, (2) phoneme or character segmentation, (3) content recognizer or pretrained encoder, (4) their training loss and procedure, and (5) the applications shown in the papers.
As can be seen and to the best of our knowledge, while there exist many controllable sequence generative models, our proposed method is the first that does not require user ID, segmentation, pretrained recognizer and proven to be applicable on both speech and handwriting domains.

\newcommand{\cmark}{\ding{108}}%
\newcommand{\hmark}{\ding{119}}%
\newcommand{\xmark}{\ding{109}}%

\begin{table}[h]
	\caption{Overview of controllable sequence models. The table provides a high-level overview of various controllable sequence models. For details, please see individual references.
	}
	\vspace{-1mm}
	\label{table: related work}
	\def\tablewidth{16mm} 
	\definecolor{mygreen}{RGB}{12, 130, 14}
	\newcommand\yes{{\color{mygreen}{\cmark $\,$}}} 
	\newcommand\ok{{\color{orange}{\hmark $\,$}}} 
	\newcommand\no{{\color{red}{\xmark $\,$ }}} 
	\centering
	\setlength{\tabcolsep}{5pt}
	\renewcommand{\arraystretch}{1.5}
	\begin{adjustbox}{max width=\textwidth}
		\begin{tabular}{p{45mm}ccccc}
			\toprule
			Method & 
			\makecell{No user ID \\ or embedding needed} & 
			\makecell{No segmentation\\ needed} &
			\makecell{No recognizer\\ needed} &
			\makecell{Training method} &
			\makecell{Domain} 
			\\
			\midrule
			\citep{kim2020glow}  &  %
			\no &  %
			\yes & %
			\yes & %
			{\color{mygreen}  log-likelihood}  &  %
			{\color{black} speech}  %
			\\
			\citep{chen2021adaspeech}  &  %
			\no &  %
			\no & %
			\yes & %
			{\color{mygreen}  log-likelihood}  &  %
			{\color{black} speech}  %
			\\
			\citep{donahue2020end}  &  %
			\no &  %
			\yes & %
			\yes & %
			{\color{black}  adversarial}  &  %
			{\color{black} speech}  %
			\\
			\citep{donahue2020end}  &  %
			\no &  %
			\yes & %
			\yes & %
			{\color{black}  adversarial}  &  %
			{\color{black} speech}  %
			\\
			\citep{kameoka2020many}  &  %
			\no &  %
			\yes & %
			\yes & %
			{\color{mygreen}  log-likelihood}  &  %
			{\color{black} speech}  %
			\\
			\citep{arik2017deep}  &  %
			\no &  %
			\ok (pretrained) & %
			\yes & %
			{\color{mygreen}  log-likelihood}  &  %
			{\color{black} speech}  %
			\\
			\citep{jia2018transfer}  &  %
			\no &  %
			\yes & %
			\yes & %
			{\color{mygreen}  log-likelihood}  &  %
			{\color{black} speech}  %
			\\
			\citep{skerry2018towards}  &  %
			\no &  %
			\yes & %
			\yes & %
			{\color{mygreen}  log-likelihood}  &  %
			{\color{black} speech}  %
			\\
			\citep{sun2020generating}  &  %
			\no &  %
			\yes & %
			\yes & %
			{\color{mygreen}  log-likelihood}  &  %
			{\color{black} speech}  %
			\\
			\midrule
			\citep{lee2021voicemixer}  &  %
			\no &  %
			\yes & %
			\yes & %
			{\color{black}  adversarial}  &  %
			{\color{black} speech}  %
			\\
			\citep{kaneko2019stargan}  &  %
			\no &  %
			\yes & %
			\yes & %
			{\color{black}  adversarial}  &  %
			{\color{black} speech}  %
			\\
			\citep{kaneko2019cyclegan}  &  %
			\no (group data)&  %
			\yes & %
			\yes & %
			{\color{black}  adversarial}  &  %
			{\color{black} speech}  %
			\\
			\citep{kaneko2018cyclegan}  &  %
			\no (group data)&  %
			\yes & %
			\yes & %
			{\color{black}  adversarial}  &  %
			{\color{black} speech}  %
			\\
			\citep{kameoka2018stargan}  &  %
			\no &  %
			\yes & %
			\yes & %
			{\color{black}  adversarial}  &  %
			{\color{black} speech}  %
			\\
			\midrule
			\citep{davis2020text}  &  %
			\yes &  %
			\ok (pretrained)  & %
			\no & %
			{\color{black}  adversarial + log-likelihood}  &  %
			{\color{black} handwriting}  %
			\\
			\citep{kang2020ganwriting}  &  %
			\no &  %
			\yes  & %
			\no & %
			{\color{black}  adversarial + log-likelihood}  &  %
			{\color{black} handwriting}  %
			\\
			\citep{bhunia2021handwriting}  &  %
			\no &  %
			\yes  & %
			\no & %
			{\color{black}  adversarial + log-likelihood}  &  %
			{\color{black} handwriting}  %
			\\
			\citep{kotani2020generating}  &  %
			\no &  %
			\ok (pretrained) & %
			\yes & %
			{\color{mygreen}   log-likelihood}  &  %
			{\color{black} handwriting}  %
			\\
			\midrule
			\citep{hsu2018hierarchical}  &  %
			\ok (not on LibriTTS)&  %
			\yes & %
			\yes & %
			{\color{mygreen}  log-likelihood}  &  %
			{\color{black} speech}  %
			\\
			\citep{hu2020unsupervised}  &  %
			\yes &  %
			\yes & %
			\ok (pretrained content encoder) & %
			\color{black}{\makecell{adversarial + log-likelihood}}  &  %
			{\color{black} speech}  %
			\\
			\citep{aksan2018deepwriting}  &  %
			\yes &  %
			\no & %
			\no & %
			\color{mygreen}{\makecell{log-likelihood}}  &  %
			{\color{black} handwriting}  %
			\\
			\citep{akuzawa2018expressive}  &  %
			\yes &  %
			\yes & %
			\yes & %
			\color{black}{log-likelihood + KL-annealing}  &  %
			{\color{black} speech}  %
			\\
			\citep{henter2018deep}  &  %
			\yes &  %
			\no & %
			\yes & %
			\color{black}{log-likelihood}  &  %
			{\color{black} speech}  %
			\\
			\citep{sun2020fully}  &  %
			\no &  %
			\no & %
			\yes & %
			\color{black}{log-likelihood}  &  %
			{\color{black} speech}  %
			\\			
			\midrule
			\citep{ma2018neural}  &  %
			\yes &  %
			\yes & %
			\yes & %
			\color{black}{adversarial}  &  %
			{\color{black} speech}  %
			\\
			\citep{graves2013generating}  &  %
			\yes &  %
			\yes & %
			\yes & %
			{\color{mygreen}  log-likelihood}  &  %
			{\color{black} handwriting}  %
			\\
			GST \citep{wang2018style}  &  %
			\yes &  %
			\yes & %
			\yes & %
			{\color{mygreen}  log-likelihood}  &  %
			{\color{black} speech}  %
			\\
			\midrule
			Proposed style equalization  &  %
			\yes &  %
			\yes & %
			\yes & %
			{\color{mygreen}  log-likelihood}  &  %
			{\color{mygreen} speech, handwriting}  %
			\\
			\bottomrule
		\end{tabular}
	\end{adjustbox}
	\vspace{-2mm}
\end{table}

\section{Additional Ablation Study: Does $\x'$ Need to be a Valid Sample?}
\label{sec: role of x'}

During training, the proposed style equalization randomly selects a sample from the training dataset as $\x'$, which is unrelated to the ground-truth $\x$. 
One interesting question can be raised naturally: ``Since $\x'$ is unrelated to $\x$, do we really need it to be a valid sample?''.
Theoretically, since our design of $\fT$ and $\fphi$ in \equref{eq: T and phi} prevents content leakage and transfers time-independent ground-truth style information from $\x$ through $\z$,  as long as $\x'$ does not contain the content information about $\x$, the learned model should be able to control style and content separately during inference. 
To verify the hypothesis, we conduct the following ablation studies: 
\begin{enumerate}[leftmargin=*]
	\item $\x'$ is a fixed vector:  We initialize $\x'$ as a random vector but fixed it during training and inference.
	\item $\x'$ is a random noise:  We randomly sample $\x'$ from the standard Gaussian distribution.
\end{enumerate}

As can be seen from Table~\ref{table: libritts noise},  both methods are able to produce high-quality models that can control content and style (low WER and \texttt{sRank} and high \texttt{cos-sim}). 
They also achieve higher \texttt{cos-sim} than GST-nS, which utilizes additional speaker information. 
Nevertheless, the proposed model (\ie, using real samples as $\x'$) is able to utilize both time-independent and \textit{time-dependent} style information of $\x$ (see discussion in \secref{sec: style bottleneck}), and thus, our model still outperforms the two models.  %

The following properties also make the proposed method more favorable than the other two cases:
\begin{itemize}[leftmargin=*]
	\item During inference, the two models still need to run $\fT$ and $\fphi$.  In comparison, our usage of $\x’$ allows us to remove $\fT$ and $\fphi$ when mimicking a reference example.
	\item Interpolation between two reference examples becomes non-trivial. By design, using a valid sample as $\x’$ enables the model to learn to transform the style from one real sample to another. 
	In contrast, the other two models only learn to transfer style to a random noise (or a fixed vector). 
\end{itemize}

\begin{table}[h]
	\caption{{Ablation study results on the role of $\x'$. All models are trained on LibriTTS-all-960 dataset.}
	}
	\vspace{-3mm}
	\label{table: libritts noise}
	\def\tablewidth{16mm} 
	\centering
	\setlength{\tabcolsep}{5pt}
	\begin{adjustbox}{max width=\textwidth}
		\begin{tabular}{p{45mm}rrrrrr}
			\toprule
			\multirow{2}[3]{*}{Method} & \multicolumn{3}{c}{Seen speakers, parallel text}  &   \multicolumn{3}{c}{Seen speakers, nonparallel text}    \\  %
			\cmidrule(lr){2-4} \cmidrule(lr){5-7} %
			& 
			\multicolumn{1}{c}{WER (\%)}  & \multicolumn{1}{c}{cos-sim $\uparrow$} & \multicolumn{1}{c}{sRank $\downarrow$} &  
			\multicolumn{1}{c}{WER (\%)}  & \multicolumn{1}{c}{cos-sim $\uparrow$} & \multicolumn{1}{c}{sRank $\downarrow$} \\ %
			\midrule
			$\x'$ is a fixed vector & 
			$8.0 \pm 0.2$ & $0.69 \pm 0.22$ & $14 \pm 78$  & 
			$7.1 \pm 0.1$ & $0.70 \pm 0.20$ & $12 \pm 95$  %
			\\
			$\x'$ is random noise & 
			$8.4 \pm 0.0$ & $0.72 \pm 0.17$ & $4.7 \pm 32$  & 
			$\mathbf{6.7 \pm 0.2}$ & $0.72 \pm 0.16$ & $5.1 \pm 56$  %
			\\
			Proposed ($\x'$ is a real sample)& 
			$\mathbf{6.2 \pm 0.5}$ & $0.82 \pm 0.14$ & $\mathbf{1.7 \pm 4.1}$  & 
			$9.4 \pm 0.3$ & $0.78 \pm 0.14$ & $\mathbf{1.8 \pm 6.0}$  %
			\\
			\midrule
			\\[0.3em]
			\multirow{2}[3]{*}{Method} & \multicolumn{3}{c}{Unseen speakers, parallel text}  &   \multicolumn{3}{c}{Unseen speakers, nonparallel text}    \\  %
			\cmidrule(lr){2-4} \cmidrule(lr){5-7} %
			& 
			\multicolumn{1}{c}{WER (\%)}  & \multicolumn{1}{c}{cos-sim $\uparrow$} & \multicolumn{1}{c}{sRank $\downarrow$} &  
			\multicolumn{1}{c}{WER (\%)}  & \multicolumn{1}{c}{cos-sim $\uparrow$} & \multicolumn{1}{c}{sRank $\downarrow$} \\ %
			\midrule
			$\x'$ is a fixed vector & 
			$9.2 \pm 0.3$ & $0.53 \pm 0.18$ & $22\pm 97$  &
			$\mathbf{6.2 \pm 0.1}$ & $0.53 \pm 0.18$ & $23\pm 103$  
			\\
			$\x'$ is random noise & 
			$8.9 \pm 0.0 $ & $0.49 \pm 0.16$ & $14\pm 55$   & 
			$6.3 \pm 0.2 $ & $0.48 \pm 0.16$ & $13\pm 56$  
			\\
			Proposed ($\x'$ is a real sample)& 
			$\mathbf{6.8 \pm 0.1}$ & $\mathbf{0.63 \pm 0.15}$ & $\mathbf{9.0\pm 63}$  & 
			$7.6 \pm 0.9$ & $0.57 \pm 0.15$ & $\mathbf{7.4\pm 43}$  
			\\
			\bottomrule
		\end{tabular}
	\end{adjustbox}
	\vspace{-0mm}
\end{table}

\section{Utilizing Time-dependent Style Information}
\label{sec: time dependency}

In the section, we try to answer the question: ``does our model utilize time-dependent style information?''
When a model fully utilizes time-dependent style information, the generated output should be a reconstruction of the reference example in the parallel setting.
On the other hand, if a model utilizes only time-independent style information, its outputs in the parallel setting can still be different from the reference example, \ie, the same sentence spoken by the same user can be very different in different contexts.
The difference in style replication quality provides the foundation to the following evaluation.

We first evaluate the capability of our model to utilize time-dependent style information by comparing the results between parallel (where all the time-dependent style information is available) and non-parallel (where little time-dependent and mostly time-independent style is available) settings.
As can be seen in Table~\ref{table: vctk}, Table~\ref{table: libritts} and \figref{figure: handwriting comparison}d, our results (WER, CER, and cos-sim) in the parallel setting are better than those in the non-parallel setting. 
The results show that the model is capable of utilizing the time-dependent information in the reference example.

\begin{figure}[h]
	\centering
	\includegraphics[width=\linewidth]{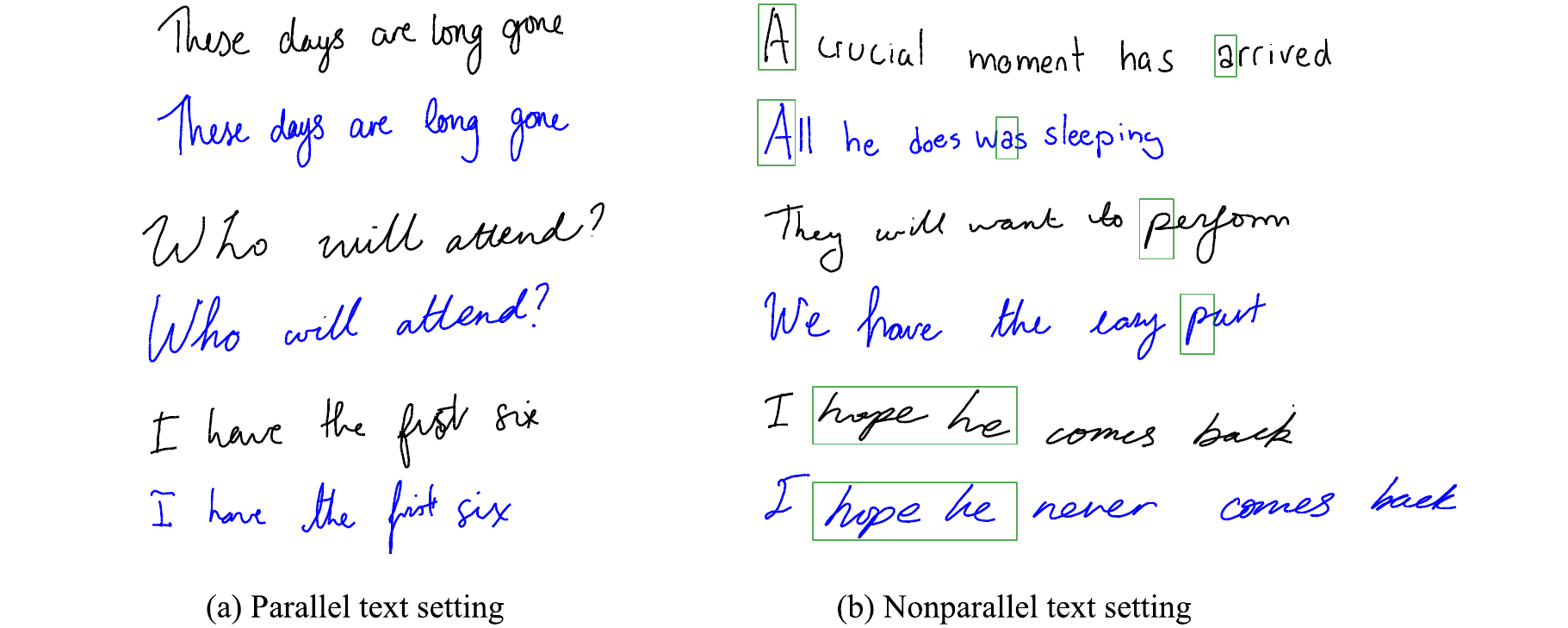}
	\caption{Parallel and nonparallel text generation results. In the figure, the reference examples are shown in black, and the generated results are shown in blue. }
	\label{fig: parallel nonparallel}
\end{figure}

Next, we verify this capability by visualizing the style attention weights in the three scenarios in \figref{figure: style attention}b and c.  
In the parallel setting (\figref{figure: style attention}b), the model strongly attends to the correct time instance, whereas in the non-parallel setting (\figref{figure: style attention}c), the model attends to localized pieces in the reference example.
Additionally, we invite the readers to qualitatively evaluate the capability via the handwriting/speech generation results provided in the supplemental website.  
As can be seen, the reproduction is “qualitatively closer“ in the parallel setting and in the nonparallel setting when a character appears in the style reference within a similar context, \eg, in \figref{fig: parallel nonparallel}b, the character ‘A’ and ‘a’ in the first non-parallel text example and ‘p’ in the second example.  
To make the observation, in \figref{fig: parallel nonparallel}b, we provide a nonparallel example that has overlapping content as the reference style.
As can be seen, the generated result is qualitatively similar to the reference example at the overlapped regions even when they are separated by unrelated text.

The ablation study in Table~\ref{table: libritts noise} compares our model (which is trained to utilize time-dependent style information via applying style equalization to 50\% of the batches) with two models that always transform style inputs (and thus can only use time-independent style information).
While all models in the comparison use the same style encoder, our model, with its capability to utilize time-dependent style information, achieves the highest cosine similarities among all three models. 

Finally, we compare the same model trained with our method (which applies style equalization to 50\% of the batches) and entirely without style equalization in \figref{figure: handwriting comparison}d.
While both models can utilize time-dependent information, the model trained without style equalization suffers from content leakage (as shown by the high CER in the nonparallel text setting).
In comparison, the proposed model is less effected by the settings.

In summary, the proposed method (utilizing the style attention module and applying style equalization on half of the batches) enables the model to utilize time-dependent style information while avoiding catastrophic content leakage.

\section{Limitations}
\label{sec: limitations}
Similar to most unsupervised methods, there is no guarantee that the proposed style equalization can \textit{entirely} prevent content leakage, \ie no ground-truth content information contained in the transformed output.
The proposed method rely on the time-invariant bottleneck to prevent the ground-truth content information leaking through the transformed style.
However, our convolutional network, which is used down-sample the input style examples, has a receptive field equal to 76 mel-spectrogram frames (about 1 second) / handwriting strokes (3 characters) in our models.
Within the receptive field, the time-dependent information is retained.
In other words, within the receptive field of the convolutional network, our assumption that we only transfer time-invariant information is violated.
While we utilize low-pass filtering before any down-sampling and avoid padding as suggested by \citet{zhang2019making}, we do not deal with the aliasing due to the nonlinearity functions (as in \cite{karras2021alias}).
Thereby, our style transformation is not entirely time-invariant and thus the transformed output can contain time-dependent information exploitable by the model.
When the model overfits the training data, our style transform can still leak content information that can eventually be utilized by the decoder. 
We observe that this phenomenon happens after the model overfits the training set, and by examining the validation loss and the KL-divergence between the prior and the posterior distributions, we can avoid it in practice.
As shown by our ablation studies comparing with/without style equalization (\figref{figure: handwriting comparison}a for handwriting and the supplemental webpage for speech), the proposed method effectively reduces content leakage.

\section{More Speech and Handwriting Generation Results}

On our website (\url{https://apple.github.io/ml-style-equalization}), we show an extensive list of speech and handwriting samples generated by the proposed and the baseline methods.
Note that it may take a while for the speech results to load, and if the audio players do not contain the play button, please increase the size of the browser window.
To remove the effect of the vocoder when comparing synthesized speech samples with real speech samples, all real speech samples (including those in the style opinion score evaluations) are converted to mel-spectrogram and reconstructed back to waveform using the same vocoder that is used by the generative models, \ie, waveglow~\citep{prenger2019waveglow}.

For speech synthesis, the webpage contains
\begin{itemize}[leftmargin=*]
	\item a video showcasing the generation of speech with various styles and content
	\item nonparallel-text generation with seen speakers from LibriTTS-all-960
	\item nonparallel-text generation with unseen speakers from LibriTTS-all-960
	\item an ablation study that compares training with and without style equalization 
	\item interpolation between two unseen style reference speech 
	\item generated speech with random styles sampled from the learned prior 
	\item nonparallel-text generation with seen speakers from VCTK
	\item parallel-text generation with seen speakers from VCTK
\end{itemize}

For handwriting synthesis, the webpage contains
\begin{itemize}[leftmargin=*]
	\item a video showcasing the online generation of handwriting with various styles and content 
	\item nonparallel-text generation with unseen style
	\item parallel-text generation with unseen style
	\item generated handwriting with random styles sampled from the learned prior 
	\item interpolation between two unseen style reference handwriting 
\end{itemize}

\end{document}